\pdfoutput=1

\documentclass[11pt]{article}
\usepackage[]{acl}
\usepackage{times}
\usepackage{latexsym}
\usepackage[T1]{fontenc}
\usepackage[utf8]{inputenc}
\usepackage[inline]{enumitem}

\usepackage{microtype}
\usepackage{inconsolata}
\usepackage{graphicx}
\usepackage{cleveref}
\usepackage{booktabs}

\usepackage{longtable}

\title{Identifying Non-Replicable Social Science Studies with Language Models}

\def\authorsep{\hspace{0.3em}}

\author{Denitsa Saynova$^{1,2*}$ 
 \authorsep Kajsa Hansson$^{3*}$  \\ 
  \textbf{Bastiaan Bruinsma}$^{1,2}$   
 \authorsep \textbf{Annika Fredén}$^{3}$
 \authorsep \textbf{Moa Johansson}$^{1,2}$ \medskip\\
\null$^{1}$ Chalmers University of Technology \quad \null$^{2}$ University of Gothenburg
\quad \null$^{3}$ Lund University\\
\texttt{\{saynova@chalmers.se, kajsa.hansson@svet.lu.se\}}}

\begin{document}
\maketitle
\def\thefootnote{*}\footnotetext{Equal contribution.}\def\thefootnote{\arabic{footnote}}
\begin{abstract}
In this study, we investigate whether LLMs can be used to indicate if a study in the behavioural social sciences is replicable. Using a dataset of 14 previously replicated studies (9 successful, 5 unsuccessful), we evaluate the ability of both open-source (Llama 3 8B, Qwen 2 7B, Mistral 7B) and proprietary (GPT-4o) instruction-tuned LLMs to discriminate between replicable and non-replicable findings. We use LLMs to generate synthetic samples of responses from behavioural studies and estimate whether the measured effects support the original findings. When compared with human replication results for these studies, we achieve F1 values of up to $77\%$ with Mistral 7B, $67\%$ with GPT-4o and Llama 3 8B, and $55\%$ with Qwen 2 7B, suggesting their potential for this task. We also analyse how effect size calculations are affected by sampling temperature and find that low variance (due to temperature) leads to biased effect estimates. 
\end{abstract}

\section{Introduction}

Since the early 2000s, several studies have shown that many findings in the behavioural social sciences are less replicable than previously thought \citep{ioannidis2005most, simmons2011false}. This has since led to a widespread debate within the community about the reliability and robustness of their research designs \citep{OSC2015a, camerer2016evaluating}. One of the first steps taken is to conduct large-scale replication projects, such as the Open Science Collaboration and Many Labs initiatives, to test the robustness of some of the more well-known studies. However, while these initiatives provide valuable insights, they are hampered by the significant financial and time demands they place on researchers, particularly early career researchers who often lack the resources to undertake such large-scale studies \citep{OpenScienceBlog2024, Whitaker2020Bropenscience}. 

\begin{figure}
    \centering
\includegraphics[width=8cm, height=6cm]{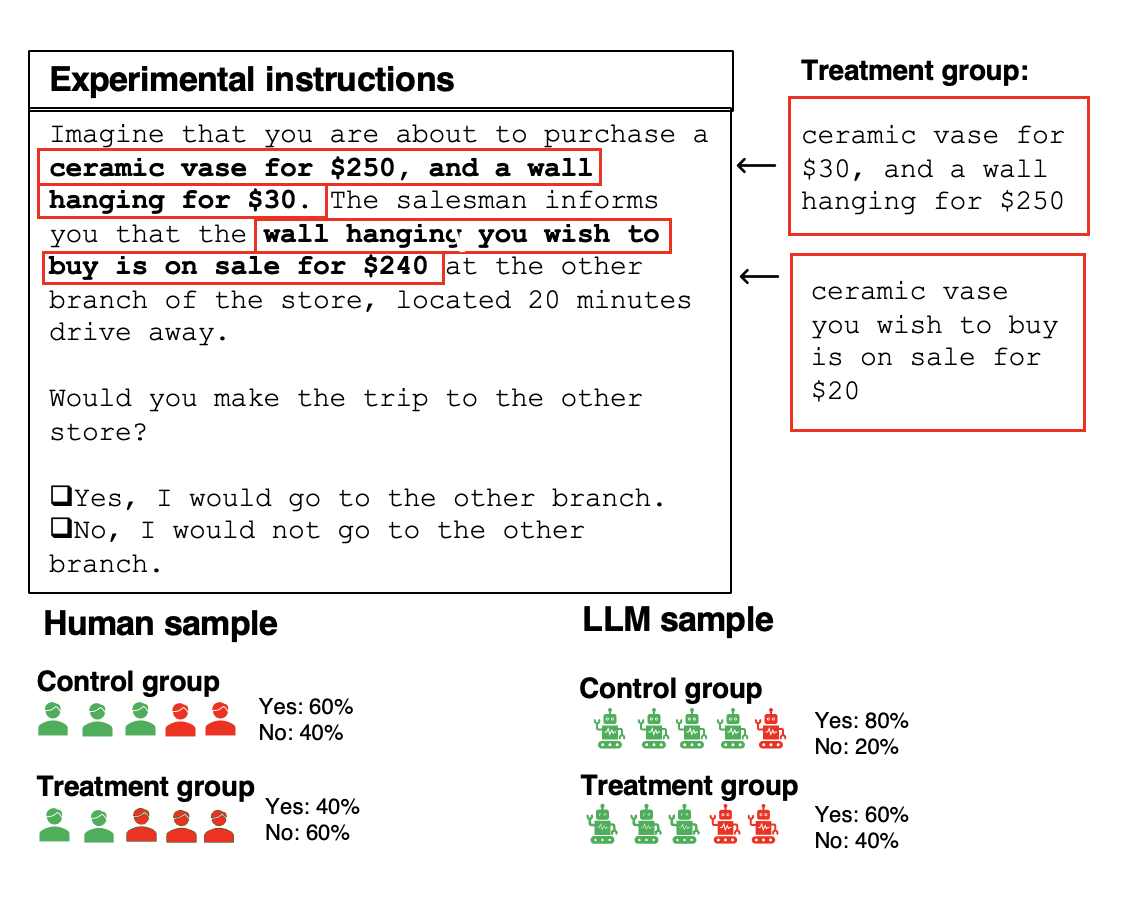}
    \caption{Illustration of a successful LLM replication (after \citet{Tversky1981a})}
    \label{fig:tversky_example}
\end{figure}

A potential way to support these efforts is to use LLMs as a ``silicon sample'' to indicate replicability of studies in the social sciences. This would be in line with the growing interest in the social sciences in using LLMs to simulate human participants in behavioural experiments \citep{park2024diminished, cui2024aireplacehumansubjects, lippert2024can, hewitt2024predicting}. And while it is clear that naively replacing humans with LLMs is not a viable approach, as LLMs suffer from problems such as hallucinations, prompt sensitivity, bias due to language, phrasing and response order \citep{gallegos2024bias,perkovic2024hallucinations,elazar2021a, Huang2025a}, there is evidence that they can capture relevant social information in some contexts, while they have difficulties in others \citep{wang2024languagemodelsservetextbased}. We do not hypothesise that LLM answer distributions capture the same distributions as observed in humans. Rather that the relative strengths of the underlying associations learned by the models may align with human behaviour enough to make them useful for this task.   

Here we propose to \emph{use LLMs to indicate whether a study should be considered for further investigation and replication efforts with human subjects}. This means that the LLM serves the purpose to indicate which studies should be looked at further, for a replication using human participants to be carried out. The LLM, then, can be used as a way of justifying why additional experiments are worth conducting.

To demonstrate our approach, we use 14 studies from the Many Labs 2 replication project that use human participants to see if using an LLM would lead to similar conclusions. A study is replicable (or successfully replicated), if the conclusion we draw from the replication aligns with the original finding. This applies when we use either human or synthetic (LLM) samples. To determine how well LLMs perform we compare the results from the LLM to the human replication. 
Previous studies have found low variance in LLM responses (including a so-called ``correct-answer effect'', where a large part of all responses fall within the same response category \citep{park2024diminished}). This can lead to an inability to detect differences between experimental conditions and biases in the statistical tests used to measure effect size. Thus, we test how temperature affects the model's ability to indicate whether a study can be successfully replicated with a human sample.

This work makes three main contributions:

\begin{itemize}[leftmargin=*, itemsep=0pt]
    \item Using samples from LLM generated responses, we find LLMs exhibit low replicability for unsuccessful human replications, while successfully replicating a proportion of the studies that have been successfully replicated with humans, suggesting their potential for predicting the replicability of a study;
    \item We find no substantial difference in performance between smaller open-source models (Llama, Qwen and Mistral) and larger proprietary models (GPT-4o);
    \item We find higher temperatures produce less usable samples (due to noise), however, lower temperatures produce a very narrow response distribution leading to zero or near-zero variation in responses and an overestimation of effect sizes;

\end{itemize}

\section{Methods}

\subsection{Data}

We take 14 studies (out of $28$) that were replicated in the Many Labs 2 replication project \citep{Klein2018a} (see \Cref{app:res_original_ml,app:summary_studies} for details). We exclude the other 14 as they contain data other than text and are therefore unsuitable for our analysis. The aim of Many Labs 2 is to confirm the results of the previous study under the same or similar conditions using (in most cases) a larger sample size. A replication is considered successful if the estimated effect is in the same \emph{direction} as in the original study and is found to be significant. Out of the 14 studies, 9 are found replicable (successfully replicated) and 5 are found non-replicable (either no significant effect or opposite effect). We use the same criterion and same statistical tests for our replications (for details see \Cref{app:stat_tests}).

An example (\Cref{fig:tversky_example}) of a study we include is the study by \citet{Tversky1981a}. In the original version, the authors presented 181 participants with a scenario in which they were buying one of two items: a relatively cheap one ($\$30$) or a more expensive one ($\$250$). Participants are asked if they are willing to travel $20$ minutes to another store for a $\$20$ discount. Both the original study and the Many Labs replication (with $7,228$ participants), found people more willing to travel when the discounted product was the cheaper item.

\subsection{Sampling}

We query LLMs with the study questions to generate synthetic samples from their responses. We use four models: three open-source: meta-llama/Meta-Llama-3-8B-Instruct (Llama), Qwen/Qwen2-7B-Instruct (Qwen), and mistral-ai/Mistral-7B-Instruct-v0.3 (Mistral), and one proprietary: gpt-4o-2024-08-06 (GPT-4o). To align with the query format of GPT-4o, we use instruction-tuned open-source models.

To generate the responses from the LLM, we follow the approach by \citet{park2024diminished}. We pass the questions in a study to the model and ask it to fill in the blanks with answers to all questions.\footnote{Code will be made available online.} We ask the LLM to return a single letter corresponding to a value on a scale (e.g. ``C'' for ``Very'') for each question. In the case of GPT-4o, we give all the questions to the model at once, whereas for the three open-source models, given their more limited ability to follow instructions, we ask the questions one by one, appending previous responses to the prompt. For each experimental setup, we produce 1000 LLM responses. These form our LLM synthetic samples.

\subsection{Temperature}

One challenge with using LLMs in this setting is that they tend to produce responses with low variance \citep{park2024diminished}. This can lead to inflated effect sizes because the standard deviations used to calculate them are very small or, if the same response is chosen for all questions, make it impossible to calculate an effect size at all. To address this, we run each LLM for $4$ different \emph{temperatures} ($3$ for GPT-4o). Temperature affects the randomness of the LLM generation and therefore the variance in its sample. The higher the temperature, the more likely the LLM is to use tokens from the low probability regions of the response distribution without changing the distribution itself. The temperatures we use are $0.1$, $0.5$, $1.0$ and $1.5$. Note that we start from $0.1$ as a temperature of $0$ would result in deterministic behaviour.

Since the increased randomness at higher temperatures can lead to a nonsensical generation, we drop a study from the analysis if this is the case. For each question the LLM is requested to choose one letter indicating value on a scale. If the LLM fails to return an allowed token for $20\%$ of the requested responses, we discard that study. We set this limit for computational reasons, as a higher percentage of responses to discard leads to higher computational costs. Note that we do not use constrained decoding, as we want to make a fair comparison to a proprietary model where we do not have access to the full distribution, and because we are interested in the effects of increased variance, and constraining the tokens may change the underlying distribution by clipping it.

We use the result from Many Labs 2 of whether the replication was successful or not as the true label of a study's replicability. Whether we find a significant effect to support the findings of the original study based on our LLM samples is used as the predicted label. As we have unbalanced data (9 successful and 5 unsuccessful replications), we measure the F1 score for each model and temperature setting, as this harmonic mean of precision and recall allows for a fairer comparison for unbalanced data. As we are primarily interested in the under-represented class of unsuccessfully replicated studies, we also calculate precision and recall.

\section{Results}

\begin{table}[ht]
\begin{tabular}{lrrrr}
\toprule
Model & F1 & $p$ & $r$ & \# Studies \\
(temperature) & & && (out of $14$)\\
\midrule
GPT-4o (0.1) & 0.50 & 0.75 & 0.38 & 11 \\
GPT-4o (0.5) & 0.67 & 0.80 & 0.57 & 10 \\
GPT-4o (1.0) & 0.36 & 0.40 & 0.33 & 10 \\
\midrule
Llama (0.1) & 0.22 & 1.0 & 0.13 & 12 \\
Llama (0.5) & 0.50 & 0.75 & 0.38 & 12 \\
Llama (1.0) & 0.67 & 1.0 & 0.50 & 11 \\
Llama (1.5) & 0.60 & 1.0 & 0.43 & 10 \\
\midrule
Qwen (0.1) & 0.46 & 0.75 & 0.33 & 13\\
Qwen (0.5) & 0.50 & 1.0 & 0.33 & 13 \\
Qwen (1.0) & 0.55 & 1.0 & 0.38 & 11 \\
Qwen (1.5) & 0.80 & 1.0 & 0.67 & 5 \\
\midrule
Mistral (0.1) & 0.36 & 1.0 & 0.22 & 13 \\
Mistral (0.5) & 0.67 & 1.0 & 0.50 & 12 \\
Mistral (1.0) & 0.77 & 0.83 & 0.71 & 11 \\
Mistral (1.5) & 0.80 & 1.0 & 0.67 & 5 \\
\bottomrule
\end{tabular}
\caption{\label{tab:temp_results} Performance metrics for each of the different temperatures and models with the number of studies replicated. High precision indicates very few of the non-replicable studies are incorrectly flagged as replicable.}
\end{table}

\Cref{tab:temp_results} shows a summary of our results. For each model and temperature setting, we have aggregated the results across all available studies. The latter number differs because the LLMs sometimes produces unusable samples. This is either because the LLM refuses to generate a response when the study dealt with 'sensitive' topics (such as in the case of \citet{inbar2009disgust}, which dealt with homosexual kissing) or the high temperature causes random tokens to be generated. In addition, especially at low temperatures, we often find extremely low variance, making it impossible to calculate an effect (these cases are treated as indicating unsuccessful replication, as a synthetic sample is still generated). 
We find LLM results align with human replications and identify non-replicable results (as shown by the high precision) and to a lesser extent successful human replications (recall). Generally models show some capability of indicating a study's replicability. Importantly, open-source models perform on par or better than the proprietary one.

\subsection{Temperature}

Our hypothesis is that increasing temperature only affects the variance of the answer distribution produced by a LLM thus allowing us to test the effects of reduced variance on the statistical tests used for calculating effect sizes and significance. By examining the histograms of answer distributions (see \Cref{app:histograms}), we find that increasing temperature indeed increases variance and does not indicate other distribution shifts.

We observe great variation of effect size magnitude estimation for both successful and unsuccessful human replicated studies (for full results see \Cref{app:eff_size}).
We find evidence to support our hypothesis that low variance (proxied by low temperature) leads to artificially high effect size magnitude. For instance, at the two lowest temperature settings -- $0.1$ and $0.5$ -- we observe 12 cases when effect sizes exceed a magnitude of $4.0$, with largest effect size of $21.6$. For comparison, no effect size in the Many Labs replication project (including original studies) exceeds an effect size of $2.0$. In most studies, and across models, we also observe a decrease of effect size magnitude with increased temperature. These two empirical results support our hypothesis that a reduced variation of sample responses may lead to biased effect size estimations. 
Furthermore, temperature settings may cause the effect size estimation to switch direction. This happens mostly for effects that are estimated to not be significant, based on p-value, but in 3 cases this causes the prediction to change from significant positive result to a significant negative effect -- for Llama and GPT-4o in \citet{risen2008people}
and for Mistral in \citet{hsee1998less}. This further exemplifies the sensitivity to this parameter.

\section{Related Work}

In previous work, \citet{park2024diminished} showed that LLMs were able to partially replicate such studies. We improve on that work in several ways. First, we extend their analysis of a single proprietary model (GPT-3.5) by comparing it to several open source models. Second, they find a ``correct response effect'' - the LLM responds in exactly the same way each time it is asked, making it impossible to calculate statistics such as effect sizes. We provide a more detailed analysis of the effect of reduced variance and find that one of the driving factors is sampling temperature.

Overall, our study extends the ways in which LLMs can be used to simulate human responses in behavioural studies. With a few exceptions \citep[cf.]{park2024diminished}, most of this literature has been concerned with simulating aspects of human behaviour by directly comparing the responses of LLMs to a sample of human participants \citep{argyle2023out, shihadeh2022brilliance, akata2023playing, aher2023using, filippas2024large, dillionlarge}, rather than focusing on how LLMs respond to experimental treatments. For example, \citet{argyle2023out} have investigated whether conditioning LLMs on socio-demographic information can generate political preferences similar to those of a representative human sample. Others have investigated whether behaviour and social preferences in economic games involving cooperation and coordination resemble those of humans \citep{akata2023playing, aher2023using, filippas2024large}, or whether GPTs make human-like moral judgments \citep{dillionlarge}. In terms of cognitive ability, \citet{strachan2024testing} compared LLMs with human performance on a battery of measures designed to measure different theories of mind. They found that GPT-4s performed at similar and sometimes higher levels than humans. An example of an experimental approach is \citet{wang2024languagemodelsservetextbased}, who find that LLMs still have difficulty replacing humans in text-based games, especially when the game has a dynamic element. This suggests that context and question type are important, which is also crucial for our experiments.

\section{Discussion and Conclusions}

Our replication of the 14 text-based experimental studies from the Many Labs project suggests that LLMs can indicate the replicability of a study with a human sample, which is promising for their potential to help social scientists select relevant studies to replicate. 
A key finding of this study is that open source models perform at a comparable level to GPT-4o in assessing the replicability of a study. In fact, GPT-4o performed slightly worse than the open source models in identifying non-replicable results. Given that previous studies exploring the potential of LLMs to simulate aspects of human behaviour have relied almost exclusively on proprietary models, this suggests that social science researchers may wish to consider open source alternatives when using LLMs as a 'silicon pattern' in their experiments.
We also consider our work to be relevant to studies of LLMs, providing insights into how to test the associations that models have encoded, and what causes them to exhibit and suppress these behaviours.

\section*{Limitations}
Our study can be seen as a first step in investigating the usefulness of LLMs in replicating social science experiments with human subjects. We also found interesting evidence that the type of experimental task matters for the usefulness of LLMs - some types of more value-laden questions led to refusals and uneven distributions that were difficult to remedy by increasing the temperature or changing the response scales. Due to the limited number of human replication studies, these results need to be further validated in a larger sample in future work. This would allow us to make stronger claims about the performance of large language models in different types of questions and environments. 
In this paper, we focus on instruction-tuned generative models in order to match the experimental setup with proprietary models and to challenge the view that they achieve better performance. These models have typically also been tuned with RLHF, which may lead to overwriting certain moral, ethical and social views present in the training data. Other types of models should be explored in future work, such as base models without tuning, encoder models, etc. Some of these may require the development of an appropriate methodology for eliciting and evaluating results. 
We focus on studies where socio-demographic characteristics are not part of the measured confounding variables (i.e. subjects are not stratified by gender, age, ethnicity, etc.). It remains an open question whether LLMs can or should be applied to studies where an effect of socio-demographic attributes is observed. 
Other ways of increasing the variance of LLMs can be explored, such as prefacing queries with random text, but this is left to future research where such approaches can be empirically validated against larger sets of studies.

\section{Ethical Considerations}
Using LLMs to predict human behaviour can lead to bias, especially when social norms and values are at stake. We therefore emphasise that our work is not intended to suggest LLMs as the sole predictors of human behaviour, but as a tool to aid replication efforts. 

\section*{Acknowledgements}

This work was partially supported by the Wallenberg AI, Autonomous Systems and Software Program -- Humanities and Society (WASP-HS) funded by the Marianne and Marcus Wallenberg Foundation and the Marcus and Amalia Wallenberg Foundation. The computations were enabled by resources provided by the National Academic Infrastructure for Supercomputing in Sweden (NAISS) at Alvis partially funded by the Swedish Research Council through grant agreement no. 2022-06725.

\bibliography{custom}

\begin{thebibliography}{29}
\providecommand{\natexlab}[1]{#1}

\bibitem[{Aher et~al.(2023)Aher, Arriaga, and Kalai}]{aher2023using}
Gati Aher, Rosa~I. Arriaga, and Adam~Tauman Kalai. 2023.
\newblock \href {https://doi.org/10.5555/3618408.3618425} {{Using Large Language Models to Simulate Multiple Humans and Replicate Human Subject Studies}}.
\newblock In \emph{International Conference on Machine Learning}, pages 337--371. PMLR, JMLR.org.

\bibitem[{Akata et~al.(2023)Akata, Schulz, Coda-Forno, Oh, Bethge, and Schulz}]{akata2023playing}
Elif Akata, Lion Schulz, Julian Coda-Forno, Seong~Joon Oh, Matthias Bethge, and Eric Schulz. 2023.
\newblock \href {https://doi.org/10.48550/ARXIV.2305.16867} {{Playing Repeated Games with Large Language Models}}.
\newblock \emph{arXiv preprint}.

\bibitem[{Argyle et~al.(2023)Argyle, Busby, Fulda, Gubler, Rytting, and Wingate}]{argyle2023out}
Lisa~P. Argyle, Ethan~C. Busby, Nancy Fulda, Joshua~R. Gubler, Christopher Rytting, and David Wingate. 2023.
\newblock \href {https://doi.org/10.1017/pan.2023.2} {{Out of One, Many: Using Language Models to Simulate Human Samples}}.
\newblock \emph{Political Analysis}, 31(3):337--351.

\bibitem[{Bahlai et~al.(2019)Bahlai, Bartlett, Burgio, Fournier, Keiser, Poisot, and Whitney}]{OpenScienceBlog2024}
Christie Bahlai, Lewis Bartlett, Kevin Burgio, Auriel Fournier, Carl Keiser, Timoth{\'{e}}e Poisot, and Kaitlin Whitney. 2019.
\newblock \href {https://doi.org/10.1511/2019.107.2.78} {{Open Science Isn't Always Open to All Scientists}}.
\newblock \emph{American Scientist}, 107(2):78.

\bibitem[{Camerer et~al.(2016)Camerer, Dreber, Forsell, Ho, Huber, Johannesson, Kirchler, Almenberg, Altmejd, Chan, Heikensten, Holzmeister, Imai, Isaksson, Nave, Pfeiffer, Razen, and Wu}]{camerer2016evaluating}
Colin~F. Camerer, Anna Dreber, Eskil Forsell, Teck-Hua Ho, J{\"{u}}rgen Huber, Magnus Johannesson, Michael Kirchler, Johan Almenberg, Adam Altmejd, Taizan Chan, Emma Heikensten, Felix Holzmeister, Taisuke Imai, Siri Isaksson, Gideon Nave, Thomas Pfeiffer, Michael Razen, and Hang Wu. 2016.
\newblock \href {https://doi.org/10.1126/science.aaf0918} {{Evaluating Replicability of Laboratory Experiments in Economics}}.
\newblock \emph{Science}, 351(6280):1433--1436.

\bibitem[{Cui et~al.(2024)Cui, Li, and Zhou}]{cui2024aireplacehumansubjects}
Ziyan Cui, Ning Li, and Huaikang Zhou. 2024.
\newblock \href {https://doi.org/10.48550/ARXIV.2409.00128} {{Can AI Replace Human Subjects? A Large-Scale Replication of Psychological Experiments with Llms}}.
\newblock \emph{arXiv preprint}.

\bibitem[{Dillion et~al.(2025)Dillion, Mondal, Tandon, and Gray}]{dillionlarge}
Danica Dillion, Debanjan Mondal, Niket Tandon, and Kurt Gray. 2025.
\newblock \href {https://doi.org/10.1038/s41598-025-86510-0} {{AI Language Model Rivals Expert Ethicist in Perceived Moral Expertise}}.
\newblock \emph{Scientific Reports}, 15(4084).

\bibitem[{Elazar et~al.(2021)Elazar, Kassner, Ravfogel, Ravichander, Hovy, Sch{\"u}tze, and Goldberg}]{elazar2021a}
Yanai Elazar, Nora Kassner, Shauli Ravfogel, Abhilasha Ravichander, Eduard Hovy, Hinrich Sch{\"u}tze, and Yoav Goldberg. 2021.
\newblock \href {https://doi.org/10.1162/tacl_a_00410} {{Measuring and Improving Consistency in Pretrained Language Models}}.
\newblock \emph{Transactions of the Association for Computational Linguistics}, 9:1012--1031.

\bibitem[{Filippas et~al.(2024)Filippas, Horton, and Manning}]{filippas2024large}
Apostolos Filippas, John~J. Horton, and Benjamin~S. Manning. 2024.
\newblock \href {https://doi.org/10.1145/3670865.3673513} {{Large Language Models as Simulated Economic Agents: What Can We Learn from Homo Silicus?}}
\newblock In \emph{Proceedings of the 25th ACM Conference on Economics and Computation}, EC '24, pages 614--615. ACM.

\bibitem[{Gallegos et~al.(2024)Gallegos, Rossi, Barrow, Tanjim, Kim, Dernoncourt, Yu, Zhang, and Ahmed}]{gallegos2024bias}
Isabel~O. Gallegos, Ryan~A. Rossi, Joe Barrow, Md~Mehrab Tanjim, Sungchul Kim, Franck Dernoncourt, Tong Yu, Ruiyi Zhang, and Nesreen~K. Ahmed. 2024.
\newblock \href {https://doi.org/10.1162/coli_a_00524} {{Bias and Fairness in Large Language Models: A Survey}}.
\newblock \emph{Computational Linguistics}, 50(3):1097--1179.

\bibitem[{Hewitt et~al.(2024)Hewitt, Ashokkumar, Ghezae, and Willer}]{hewitt2024predicting}
Luke Hewitt, Ashwini Ashokkumar, Isaias Ghezae, and Robb Willer. 2024.
\newblock {Predicting Results of Social Science Experiments Using Large Language Models}.
\newblock Https://docsend.com/view/ity6yf2dansesucf.

\bibitem[{Hsee(1998)}]{hsee1998less}
Christopher~K. Hsee. 1998.
\newblock \href {https://doi.org/10.1002/(sici)1099-0771(199806)11:2<107::aid-bdm292>3.0.co;2-y} {{Less Is Better: When Low-Value Options Are Valued More Highly Than High-Value Options}}.
\newblock \emph{Journal of Behavioral Decision Making}, 11(2):107--121.

\bibitem[{Huang et~al.(2025)Huang, Yu, Ma, Zhong, Feng, Wang, Chen, Peng, Feng, Qin, and Liu}]{Huang2025a}
Lei Huang, Weijiang Yu, Weitao Ma, Weihong Zhong, Zhangyin Feng, Haotian Wang, Qianglong Chen, Weihua Peng, Xiaocheng Feng, Bing Qin, and Ting Liu. 2025.
\newblock \href {https://doi.org/10.1145/3703155} {{A Survey on Hallucination in Large Language Models: Principles, Taxonomy, Challenges, and Open Questions}}.
\newblock \emph{ACM Transactions on Information Systems}, 43(2):1--55.

\bibitem[{Inbar et~al.(2009)Inbar, Pizarro, Knobe, and Bloom}]{inbar2009disgust}
Yoel Inbar, David~A. Pizarro, Joshua Knobe, and Paul Bloom. 2009.
\newblock \href {https://doi.org/10.1037/a0015960} {{Disgust Sensitivity Predicts Intuitive Disapproval of Gays}}.
\newblock \emph{Emotion}, 9(3):435--439.

\bibitem[{Ioannidis(2005)}]{ioannidis2005most}
John P.~A. Ioannidis. 2005.
\newblock \href {https://doi.org/10.1371/journal.pmed.0020124} {{Why Most Published Research Findings Are False}}.
\newblock \emph{PLoS Medicine}, 2(8):e124.

\bibitem[{Klein et~al.(2018)Klein, Vianello, Hasselman, Adams, Adams, Alper, Aveyard, Axt, Babalola, Bahn{\'{i}}k, Batra, Berkics, Bernstein, Berry, Bialobrzeska, Binan, Bocian, Brandt, Busching, R{\'{e}}dei, Cai, Cambier, Cantarero, Carmichael, Ceric, Chandler, Chang, Chatard, Chen, Cheong, Cicero, Coen, Coleman, Collisson, Conway, Corker, Curran, Cushman, Dagona, Dalgar, Dalla~Rosa, Davis, de~Bruijn, De~Schutter, Devos, de~Vries, Do{\u{g}}ulu, Dozo, Dukes, Dunham, Durrheim, Ebersole, Edlund, Eller, English, Finck, Frankowska, Freyre, Friedman, Galliani, Gandi, Ghoshal, Giessner, Gill, Gnambs, G{\'{o}}mez, Gonz{\'{a}}lez, Graham, Grahe, Grahek, Green, Hai, Haigh, Haines, Hall, Heffernan, Hicks, Houdek, Huntsinger, Huynh, IJzerman, Inbar, Innes-Ker, Jim{\'{e}}nez-Leal, John, Joy-Gaba, Kamilo{\u{g}}lu, Kappes, Karabati, Karick, Keller, Kende, Kervyn, Kne{\v{z}}evi{\'{c}}, Kovacs, Krueger, Kurapov, Kurtz, Lakens, Lazarevi{\'{c}}, Levitan, Lewis, Lins, Lipsey, Losee, Maassen, Maitner, Malingumu, Mallett,
  Marotta, Me{{\dj}}edovi{\'{c}}, Mena-Pacheco, Milfont, Morris, Murphy, Myachykov, Neave, Neijenhuijs, Nelson, Neto, Lee~Nichols, Ocampo, O{\textquoteright}Donnell, Oikawa, Oikawa, Ong, Orosz, Osowiecka, Packard, P{\'{e}}rez-S{\'{a}}nchez, Petrovi{\'{c}}, Pilati, Pinter, Podesta, Pogge, Pollmann, Rutchick, Saavedra, Saeri, Salomon, Schmidt, Sch{\"{o}}nbrodt, Sekerdej, Sirlop{\'{u}}, Skorinko, Smith, Smith-Castro, Smolders, Sobkow, Sowden, Spachtholz, Srivastava, Steiner, Stouten, Street, Sundfelt, Szeto, Szumowska, Tang, Tanzer, Tear, Theriault, Thomae, Torres, Traczyk, Tybur, Ujhelyi, van Aert, van Assen, van~der Hulst, van Lange, van {\textquoteright}t~Veer, V{\'{a}}squez-Echeverr{\'{i}}a, Ann~Vaughn, V{\'{a}}zquez, Vega, Verniers, Verschoor, Voermans, Vranka, Welch, Wichman, Williams, Wood, Woodzicka, Wronska, Young, Zelenski, Zhijia, and Nosek}]{Klein2018a}
Richard~A. Klein, Michelangelo Vianello, Fred Hasselman, Byron~G. Adams, Reginald~B. Adams, Sinan Alper, Mark Aveyard, Jordan~R. Axt, Mayowa~T. Babalola, {{\v{S}}}t{\v{e}}p{\'{a}}n Bahn{\'{i}}k, Rishtee Batra, Mih{\'{a}}ly Berkics, Michael~J. Bernstein, Daniel~R. Berry, Olga Bialobrzeska, Evans~Dami Binan, Konrad Bocian, Mark~J. Brandt, Robert Busching, Anna~Cabak R{\'{e}}dei, Huajian Cai, Fanny Cambier, Katarzyna Cantarero, Cheryl~L. Carmichael, Francisco Ceric, Jesse Chandler, Jen-Ho Chang, Armand Chatard, Eva~E. Chen, Winnee Cheong, David~C. Cicero, Sharon Coen, Jennifer~A. Coleman, Brian Collisson, Morgan~A. Conway, Katherine~S. Corker, Paul~G. Curran, Fiery Cushman, Zubairu~K. Dagona, Ilker Dalgar, Anna Dalla~Rosa, William~E. Davis, Maaike de~Bruijn, Leander De~Schutter, Thierry Devos, Marieke de~Vries, Canay Do{\u{g}}ulu, Nerisa Dozo, Kristin~Nicole Dukes, Yarrow Dunham, Kevin Durrheim, Charles~R. Ebersole, John~E. Edlund, Anja Eller, Alexander~Scott English, Carolyn Finck, Natalia Frankowska,
  Miguel-{{\'{A}}}ngel Freyre, Mike Friedman, Elisa~Maria Galliani, Joshua~C. Gandi, Tanuka Ghoshal, Steffen~R. Giessner, Tripat Gill, Timo Gnambs, {{\'{A}}}ngel G{\'{o}}mez, Roberto Gonz{\'{a}}lez, Jesse Graham, Jon~E. Grahe, Ivan Grahek, Eva G.~T. Green, Kakul Hai, Matthew Haigh, Elizabeth~L. Haines, Michael~P. Hall, Marie~E. Heffernan, Joshua~A. Hicks, Petr Houdek, Jeffrey~R. Huntsinger, Ho~Phi Huynh, Hans IJzerman, Yoel Inbar, {{\AA}}se~H. Innes-Ker, William Jim{\'{e}}nez-Leal, Melissa-Sue John, Jennifer~A. Joy-Gaba, Roza~G. Kamilo{\u{g}}lu, Heather~Barry Kappes, Serdar Karabati, Haruna Karick, Victor~N. Keller, Anna Kende, Nicolas Kervyn, Goran Kne{\v{z}}evi{\'{c}}, Carrie Kovacs, Lacy~E. Krueger, German Kurapov, Jamie Kurtz, Dani{\"{e}}l Lakens, Ljiljana~B. Lazarevi{\'{c}}, Carmel~A. Levitan, Neil~A. Lewis, Samuel Lins, Nikolette~P. Lipsey, Joy~E. Losee, Esther Maassen, Angela~T. Maitner, Winfrida Malingumu, Robyn~K. Mallett, Satia~A. Marotta, Janko Me{{\dj}}edovi{\'{c}}, Fernando Mena-Pacheco,
  Taciano~L. Milfont, Wendy~L. Morris, Sean~C. Murphy, Andriy Myachykov, Nick Neave, Koen Neijenhuijs, Anthony~J. Nelson, F{\'{e}}lix Neto, Austin Lee~Nichols, Aaron Ocampo, Susan~L. O{\textquoteright}Donnell, Haruka Oikawa, Masanori Oikawa, Elsie Ong, G{\'{a}}bor Orosz, Malgorzata Osowiecka, Grant Packard, Rolando P{\'{e}}rez-S{\'{a}}nchez, Boban Petrovi{\'{c}}, Ronaldo Pilati, Brad Pinter, Lysandra Podesta, Gabrielle Pogge, Monique M.~H. Pollmann, Abraham~M. Rutchick, Patricio Saavedra, Alexander~K. Saeri, Erika Salomon, Kathleen Schmidt, Felix~D. Sch{\"{o}}nbrodt, Maciej~B. Sekerdej, David Sirlop{\'{u}}, Jeanine L.~M. Skorinko, Michael~A. Smith, Vanessa Smith-Castro, Karin C. H.~J. Smolders, Agata Sobkow, Walter Sowden, Philipp Spachtholz, Manini Srivastava, Troy~G. Steiner, Jeroen Stouten, Chris N.~H. Street, Oskar~K. Sundfelt, Stephanie Szeto, Ewa Szumowska, Andrew C.~W. Tang, Norbert Tanzer, Morgan~J. Tear, Jordan Theriault, Manuela Thomae, David Torres, Jakub Traczyk, Joshua~M. Tybur, Adrienn Ujhelyi,
  Robbie C.~M. van Aert, Marcel A. L.~M. van Assen, Marije van~der Hulst, Paul A.~M. van Lange, Anna~Elisabeth van {\textquoteright}t~Veer, Alejandro V{\'{a}}squez-Echeverr{\'{i}}a, Leigh Ann~Vaughn, Alexandra V{\'{a}}zquez, Luis~Diego Vega, Catherine Verniers, Mark Verschoor, Ingrid P.~J. Voermans, Marek~A. Vranka, Cheryl Welch, Aaron~L. Wichman, Lisa~A. Williams, Michael Wood, Julie~A. Woodzicka, Marta~K. Wronska, Liane Young, John~M. Zelenski, Zeng Zhijia, and Brian~A. Nosek. 2018.
\newblock \href {https://doi.org/10.1177/2515245918810225} {{Many Labs 2: Investigating Variation in Replicability across Samples and Settings}}.
\newblock \emph{Advances in Methods and Practices in Psychological Science}, 1(4):443--490.

\bibitem[{Lippert et~al.(2024)Lippert, Dreber, Johannesson, Tierney, Cyrus-Lai, Uhlmann, {Emotion Expression Collaboration}, and Pfeiffer}]{lippert2024can}
Steffen Lippert, Anna Dreber, Magnus Johannesson, Warren Tierney, Wilson Cyrus-Lai, Eric~Luis Uhlmann, {Emotion Expression Collaboration}, and Thomas Pfeiffer. 2024.
\newblock \href {https://doi.org/10.1098/rsos.240682} {{Can Large Language Models Help Predict Results from a Complex Behavioural Science Study?}}
\newblock \emph{Royal Society Open Science}, 11(9):240682.

\bibitem[{{Open Science Collaboration}(2015)}]{OSC2015a}
{Open Science Collaboration}. 2015.
\newblock \href {https://doi.org/10.1126/science.aac4716} {{Estimating the Reproducibility of Psychological Science}}.
\newblock \emph{Science}, 349(6251):aac4716.

\bibitem[{Park et~al.(2024)Park, Schoenegger, and Zhu}]{park2024diminished}
Peter~S. Park, Philipp Schoenegger, and Chongyang Zhu. 2024.
\newblock \href {https://doi.org/10.3758/s13428-023-02307-x} {{Diminished Diversity-Of-Thought in a Standard Large Language Model}}.
\newblock \emph{Behavior Research Methods}, 56:5754--5770.

\bibitem[{Perkovi{\'c} et~al.(2024)Perkovi{\'c}, Drobnjak, and Boti{\v{c}}ki}]{perkovic2024hallucinations}
Gabrijela Perkovi{\'c}, Antun Drobnjak, and Ivica Boti{\v{c}}ki. 2024.
\newblock \href {https://doi.org/10.1109/mipro60963.2024.10569238} {{Hallucinations in LLMs: Understanding and Addressing Challenges}}.
\newblock In \emph{2024 47th MIPRO ICT and Electronics Convention (MIPRO)}, pages 2084--2088. IEEE, IEEE.

\bibitem[{Risen and Gilovich(2008)}]{risen2008people}
Jane~L. Risen and Thomas Gilovich. 2008.
\newblock \href {https://doi.org/10.1037/0022-3514.95.2.293} {{Why People Are Reluctant to Tempt Fate}}.
\newblock \emph{Journal of Personality and Social Psychology}, 95(2):293--307.

\bibitem[{Schwarz et~al.(1991)Schwarz, Strack, and Mai}]{SCHWARZ}
Norbert Schwarz, Fritz Strack, and Hans-Peter Mai. 1991.
\newblock \href {https://doi.org/10.1086/269239} {{Assimilation and Contrast Effects in Part-Whole Question Sequences: A Conversational Logic Analysis}}.
\newblock \emph{Public Opinion Quarterly}, 55(1):3--23.

\bibitem[{Shafir(1993)}]{shafir1993choosing}
Eldar Shafir. 1993.
\newblock \href {https://doi.org/10.3758/bf03197186} {{Choosing Versus Rejecting: Why Some Options Are Both Better and Worse Than Others}}.
\newblock \emph{Memory \& Cognition}, 21(4):546--556.

\bibitem[{Shihadeh et~al.(2022)Shihadeh, Ackerman, Troske, Lawson, and Gonzalez}]{shihadeh2022brilliance}
Juliana Shihadeh, Margareta Ackerman, Ashley Troske, Nicole Lawson, and Edith Gonzalez. 2022.
\newblock \href {https://doi.org/10.1109/ghtc55712.2022.9910995} {{Brilliance Bias in GPT-3}}.
\newblock In \emph{2022 IEEE Global Humanitarian Technology Conference (GHTC)}, pages 62--69. IEEE, IEEE.

\bibitem[{Simmons et~al.(2011)Simmons, Nelson, and Simonsohn}]{simmons2011false}
Joseph~P. Simmons, Leif~D. Nelson, and Uri Simonsohn. 2011.
\newblock \href {https://doi.org/10.1177/0956797611417632} {{False-Positive Psychology: Undisclosed Flexibility in Data Collection and Analysis Allows Presenting Anything as Significant}}.
\newblock \emph{Psychological Science}, 22(11):1359--1366.

\bibitem[{Strachan et~al.(2024)Strachan, Albergo, Borghini, Pansardi, Scaliti, Gupta, Saxena, Rufo, Panzeri, Manzi, Graziano, and Becchio}]{strachan2024testing}
James W.~A. Strachan, Dalila Albergo, Giulia Borghini, Oriana Pansardi, Eugenio Scaliti, Saurabh Gupta, Krati Saxena, Alessandro Rufo, Stefano Panzeri, Guido Manzi, Michael S.~A. Graziano, and Cristina Becchio. 2024.
\newblock \href {https://doi.org/10.1038/s41562-024-01882-z} {{Testing Theory of Mind in Large Language Models and Humans}}.
\newblock \emph{Nature Human Behaviour}, 8(7):1285--1295.

\bibitem[{Tversky and Kahneman(1981)}]{Tversky1981a}
Amos Tversky and Daniel Kahneman. 1981.
\newblock \href {https://doi.org/10.1126/science.7455683} {{The Framing of Decisions and the Psychology of Choice}}.
\newblock \emph{Science}, 211(4481):453--458.

\bibitem[{Wang et~al.(2024)Wang, Todd, Xiao, Yuan, C{\^{o}}t{\'{e}}, Clark, and Jansen}]{wang2024languagemodelsservetextbased}
Ruoyao Wang, Graham Todd, Ziang Xiao, Xingdi Yuan, Marc-Alexandre C{\^{o}}t{\'{e}}, Peter Clark, and Peter Jansen. 2024.
\newblock \href {https://doi.org/10.48550/arxiv.2406.06485} {{Can Language Models Serve As Text-Based World Simulators?}}

\bibitem[{Whitaker and Guest(2020)}]{Whitaker2020Bropenscience}
Kirstie Whitaker and Olivia Guest. 2020.
\newblock {\#bropenscience is broken science: Kirstie Whitaker and Olivia Guest ask how open ``open science'' really is.}
\newblock \emph{The Psychologist}, 33:34--37.

\end{thebibliography}

\clearpage
\appendix

\section{Replicability of results}
\label{app:res_original_ml}

The effects measured in the original study and in the human replication can be seen in \Cref{tab:appendix_a}. If conclusions agree, study is seen as replicable.

\begin{table}[ht]
\small
\begin{tabular}{llll}
\toprule
\textbf{\#} & \textbf{Study}        & \textbf{Original} & \textbf{Many Labs} \\ 
\midrule
2                 & Kay (2013)                   & Positive                 & Non-significant    \\
4                 & Graham et al. (2009)         & Positive                 & Positive           \\
5                 & Rottenstreich \& Hsee (2001) & Positive                 & Negative           \\
6                 & Bauer et al. (2012)          & Positive                 & Positive           \\
8                 & Inbar et al. (2009)          & Positive                 & Non-significant    \\
11                & Hauser et al. (2007)         & Positive                 & Positive           \\
13                & Ross et al. (1977a)          & Positive                 & Positive           \\
14                & Ross et al. (1977b)          & Positive                 & Positive           \\
16                & Tversky \& Kahneman (1981)   & Positive                 & Positive           \\
18                & Risen \& Gilovich (2008)     & Positive                 & Positive           \\
21                & Hsee (1998)                  & Positive                 & Positive           \\
24                & Schwarz et al. (1991)        & Positive                 & Non-significant    \\
25                & Shafir (1993)                & Positive                 & Negative           \\
27                & Knobe (2003)                 & Positive                 & Positive           \\ 
\bottomrule
\end{tabular}
\caption{Summary of the effects found for both the original study and the Many Labs 2 replication.}
\label{tab:appendix_a} 
\end{table}

\section{Summary of studies}
\label{app:summary_studies}

Text-based studies that have been replicated with human samples by the Many Labs (ML) project, including description, result, and replicability can be seen in \Cref{tab:overview}.

\clearpage
\onecolumn

\begin{longtable}{lp{2cm}p{4cm}p{4cm}p{3cm}}
\caption{Overview of the 14 studies taken from Many Labs 2\label{tab:overview}}\\
\toprule
 \textbf{\#} & \textbf{Authors} & \textbf{Description} & \textbf{Original Result} & \textbf{ML Replicable?} \\
 \midrule
\endfirsthead
\multicolumn{5}{r@{}}{\Cref{tab:overview} continued from the previous page}\\
\toprule
 \textbf{\#} & \textbf{Authors} & \textbf{Description} & \textbf{Original Result} & \textbf{ML Replicable?} \\
 \midrule
\endhead 
\bottomrule
\endfoot
\endlastfoot
2. & Kay et al., 2014 & Subjects read a passage in which a natural event was described as either structured or random. The effect on subjects' willingness to pursue their goal was measured & Structured events are associated with higher willingness to pursue goals & No \\
4. & Graham et al. 2009 & Subjects first self-identified on the liberal-conservative spectrum. The effect of this on whether in-group, authority, or purity concerns were considered more relevant to moral judgments was measured & Individuals on the political left prioritize harm and fairness in moral judgments, whereas those on the political right emphasize in-group loyalty, authority, and purity & Yes \\
5. & Rottenstreich \& Hsee, 2001 & Subjects were asked to choose between a kiss from a favourite movie star and \$50, either with a certain outcome or with only a 1\% chance of getting the outcome (affective vs. financially attractive options) & Participants preferred the kiss more when the outcome was unlikely compared to when the outcome was certain. & No. Opposite direction. \\
6. & Bauer et al., 2012 & Subjects read a passage in which they and others were described as either "consumers" or "individuals". The effect on whether they trusted others to conserve water was measured & Consumer framing led to lower trust & Yes \\
8. & Inbar et al., 2009 & Subjects read a passage about a director encouraging homosexual versus heterosexual kissing. The effect of their disgust sensitivity on whether they thought the encouragement was intentional was measured & Actions are seen as more intentional for homosexual kissing than for heterosexual kissing & No \\
11. & Hauser et al., 2007 & Subjects were asked whether they would sacrifice one life to save five, either by changing the path of a runaway trolley or by pushing a large man in front of a trolley & When it came to changing the path of a runaway trolley, 89\% of participants thought the action was permissible, but when it came to saving five lives by pushing a large man, only 11\% thought it was permissible & Yes \\
13. & Ross et al., 1977 & In a supermarket scenario, subjects estimated whether they and others would sign a release for a TV commercial. Their estimated probability of signing and the probability of others signing were compared. & Choosing an option is associated with a higher estimate of the frequency of that choice. & Yes \\
14. & Ross et al., 1977 & In a traffic ticket scenario, subjects estimated whether they and others would either pay the fine or go to court. Their estimated probability of paying the fine was compared with the probability of others & Choosing an option is associated with a higher estimate of the frequency of that choice & Yes \\
16. & Tversky \& Kahneman, 1981 & In a scenario involving the purchase of a cheap item and an expensive item, subjects answered whether they would buy at a distant store with a fixed discount on either the cheap item or the expensive item & Participants were more likely to say they would go to the other store if the cheap item was on sale than if the expensive item was on sale & Yes \\
 18. & Risen \& Gilovich, 2008 & In the role of a student in class, subjects estimated the likelihood of being called on when they were told they had not prepared for class (tempting fate) versus when they were told they had prepared (not tempting fate) & Likelihood estimates were higher when fate was tempting & Yes \\
21. & Hsee, 1998 & Subjects estimated the degree of generosity of a cheaper gift in a more expensive category versus a more expensive gift in a less expensive category & The higher-priced cheaper item is perceived as more generous than the lower-priced expensive item & Yes \\
24. & Schwarz et al., 1991 & Subjects were asked "How satisfied are you with your relationship?" and "How satisfied are you with your life as a whole?" in the two possible orders & Asking specific life satisfaction questions before general ones resulted in higher correlations & No \\
25. & Shafir, 1993 & Subjects chose whether to award or deny custody to one of two parents: a parent with extreme characteristics (either very positive or very negative) and a parent with average characteristics & The extreme parent was more likely to be awarded custody and more likely to be denied custody. (Negative traits were weighted more heavily than positive traits in rejecting options, and positive traits were weighted more heavily than negative traits in selecting options) & No. The results are in the opposite direction. \\
27. & Knobe, 2003 & Subjects were asked whether a company vice-president's decision to produce a helpful or harmful side effect was intentional. The two were compared & Subjects were more likely to believe the agent acted intentionally if the side effect was harmful rather than helpful. & Yes \\ 
\bottomrule
\end{longtable}

\clearpage
\twocolumn

\newpage

\section{Statistical tests used in Many Labs}
\label{app:stat_tests}

All studies use Cohen's d for measuring effect size, apart from \citet{Tversky1981a}, which uses odds ratio and \citet{SCHWARZ}, which uses Cohen's q. Significance (p-value) is measured with a t-test for all studies apart from \citet{Tversky1981a} (chi-squared test), \citet{SCHWARZ} (z Fisher test), and \citet{shafir1993choosing} (z test).

\section{Histograms of answer responses}
\label{app:histograms}

\begin{figure*}[!tb]
\caption{Histograms of answer distributions for open-source models compared to human sample and GPT-3.5 sample \citep{park2024diminished}}
\includegraphics[width=\linewidth]{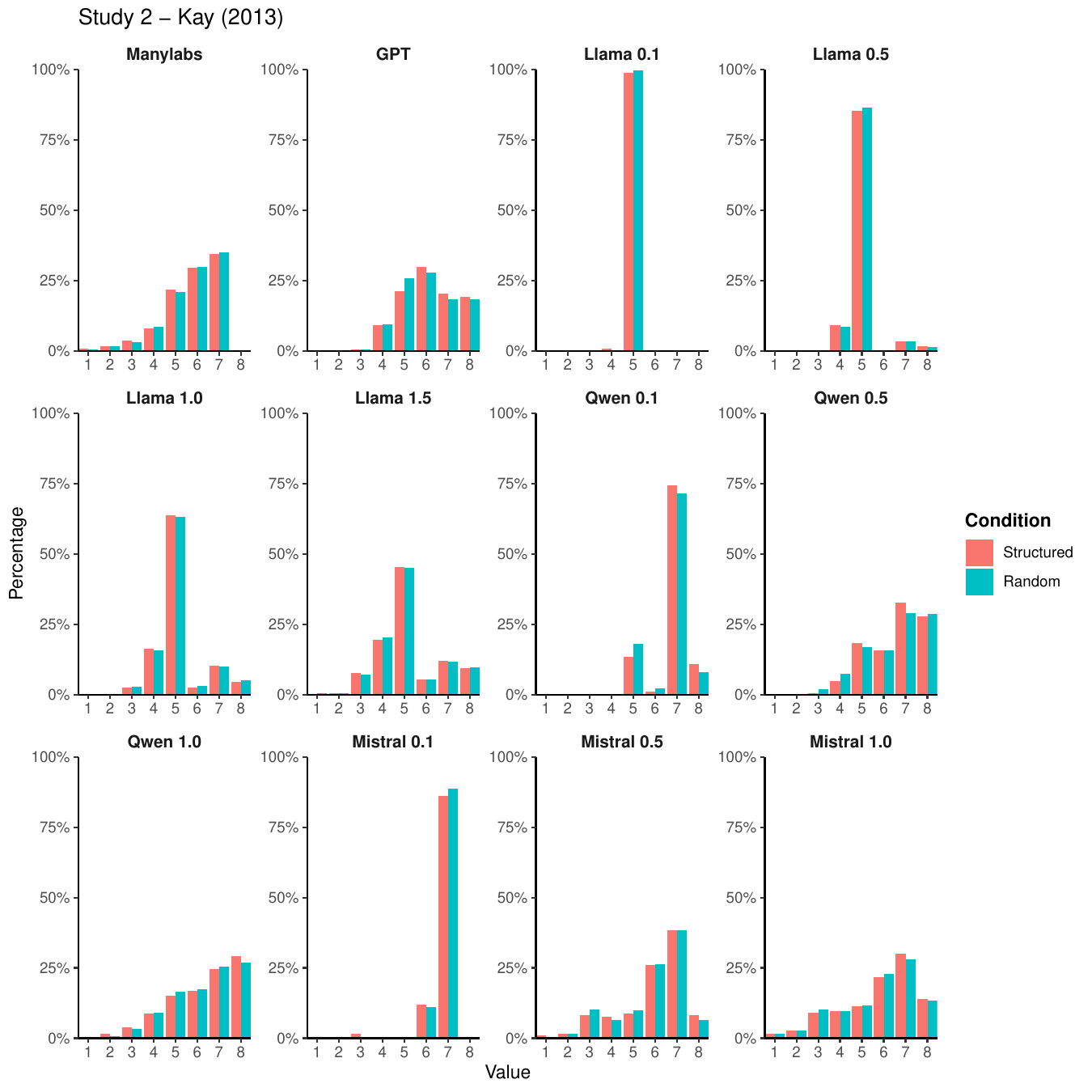}
\label{fig:appendixb_study2} 
\end{figure*}

Histograms of the answer distribution of the open-source models can be seen in \Cref{fig:appendixb_study2}. We also include the human sample distribution and a previous LLM replication study based on GPT-3.5 \cite{park2024diminished}.

\begin{figure*}[!tb]
\includegraphics[width=\linewidth]{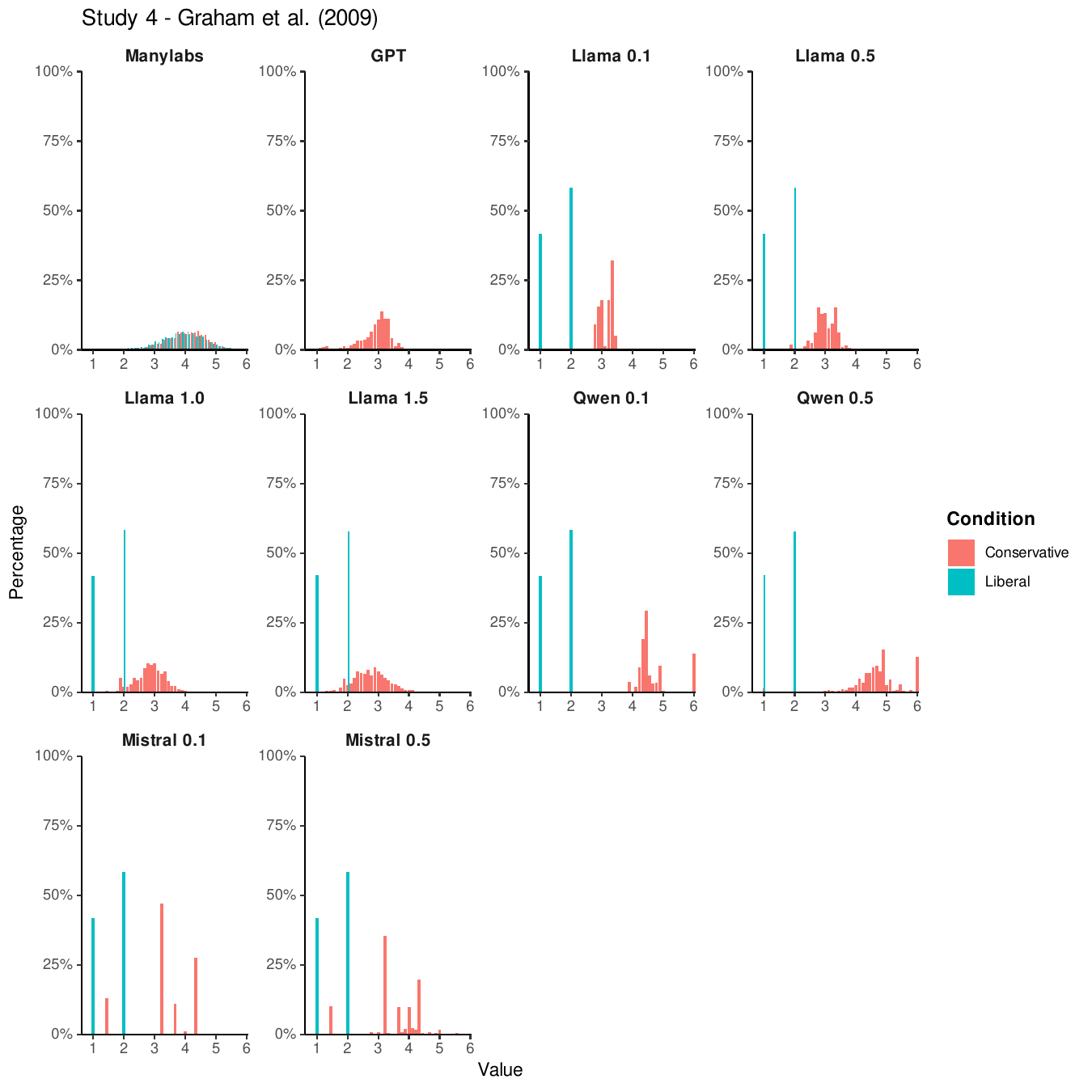}
\end{figure*}

\begin{figure*}[!tb]
\includegraphics[width=\linewidth]{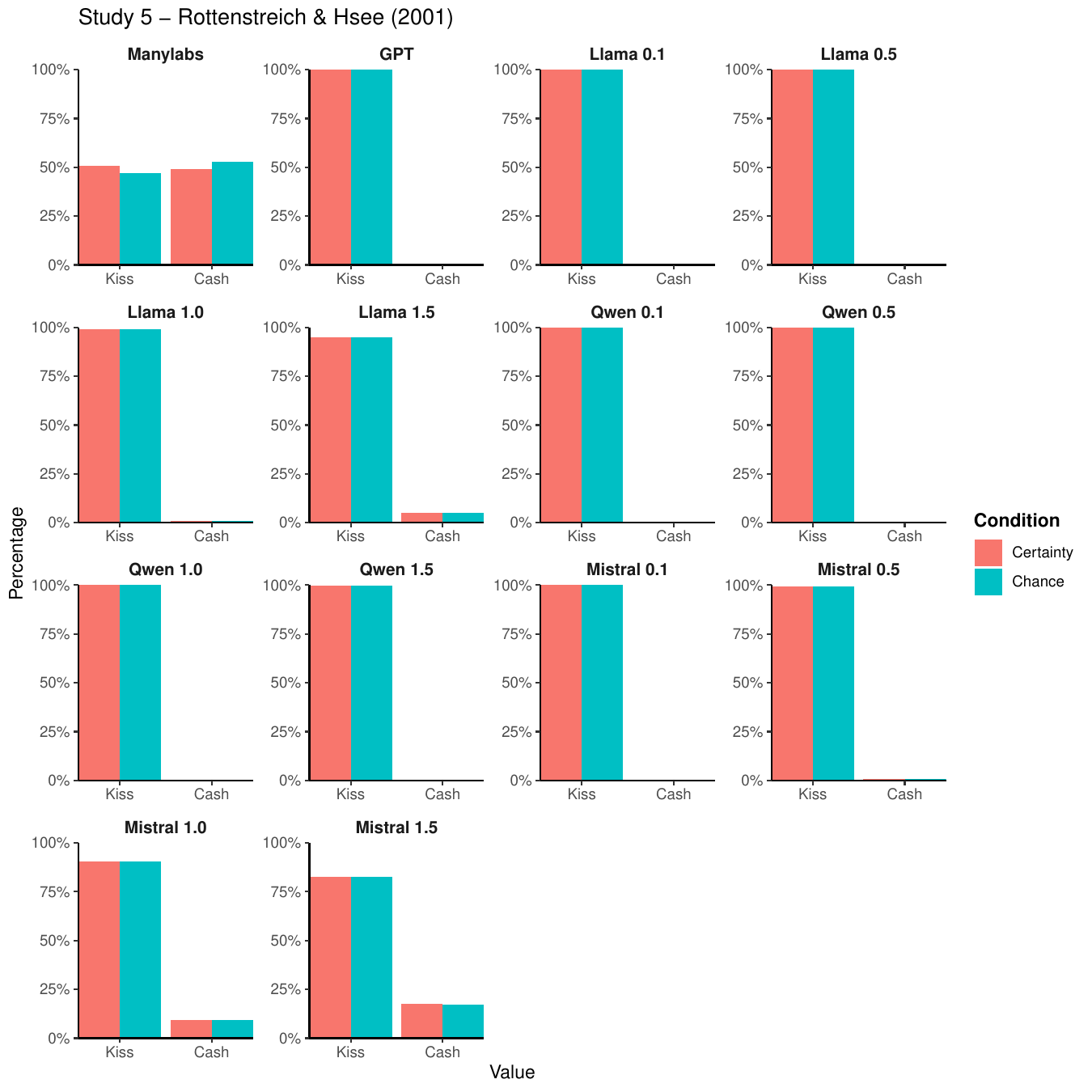}
\end{figure*}

\begin{figure*}[!tb]
\includegraphics[width=\linewidth]{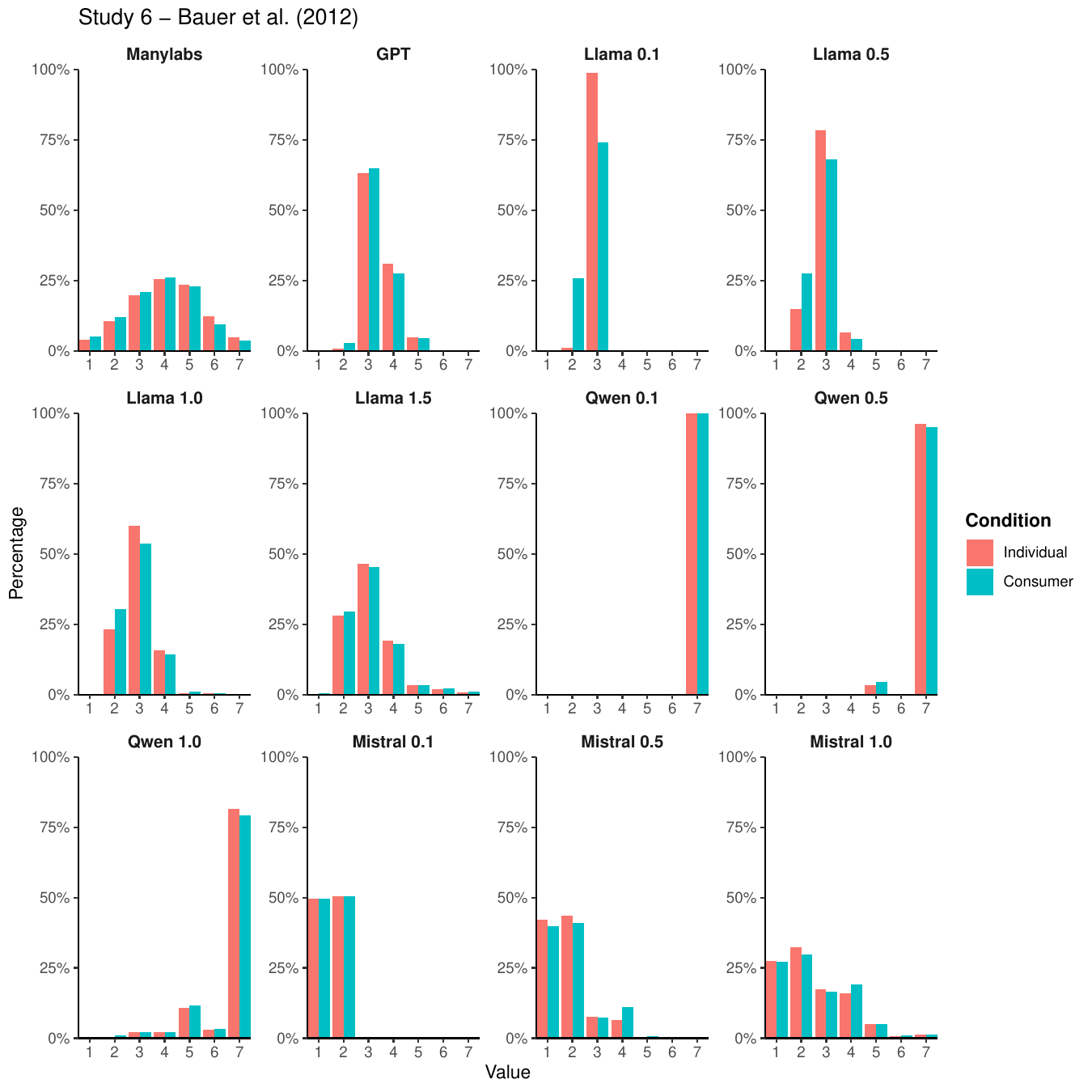}
\end{figure*}

\begin{figure*}[!tb]
\includegraphics[width=\linewidth]{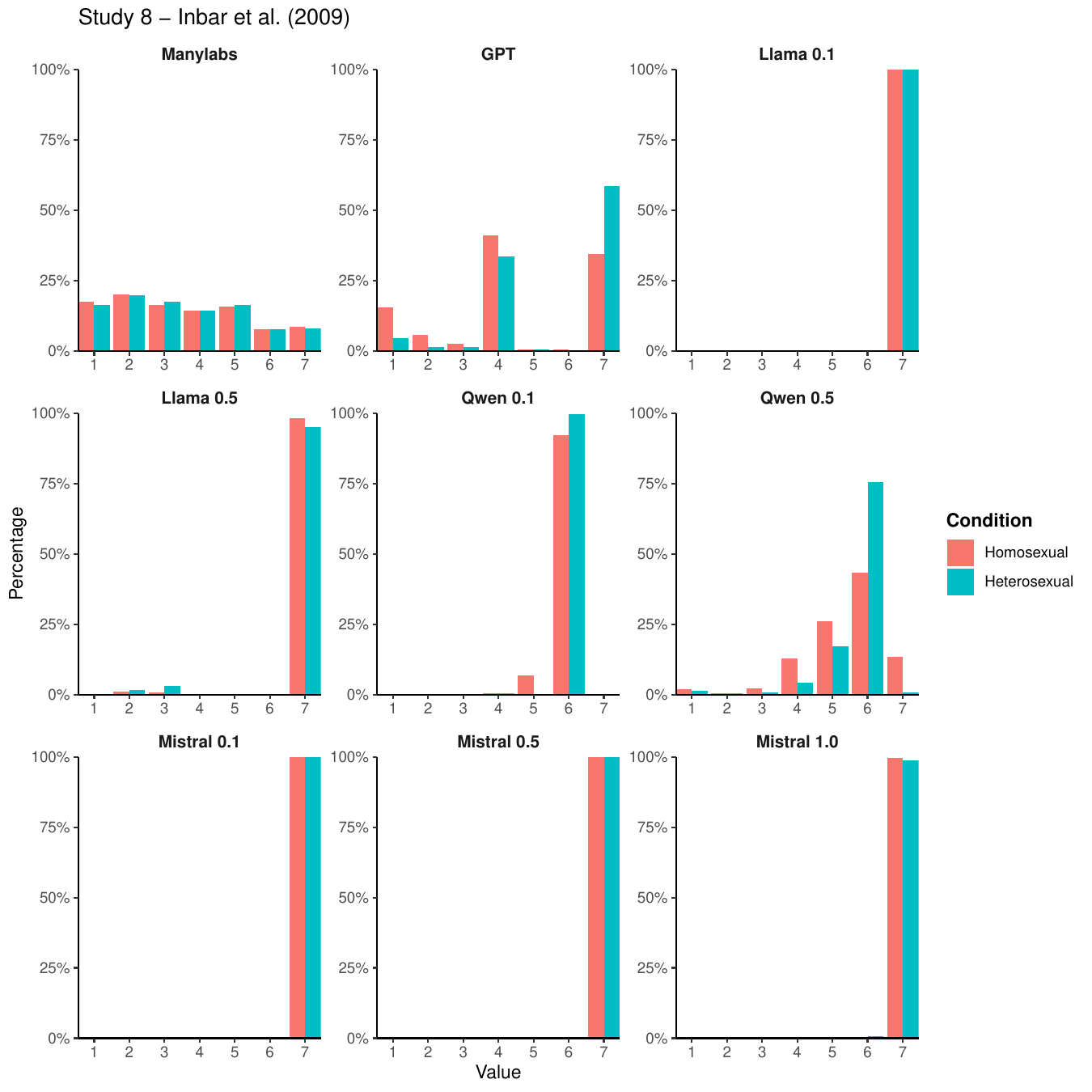}
\end{figure*}

\begin{figure*}[!tb]
\includegraphics[width=\linewidth]{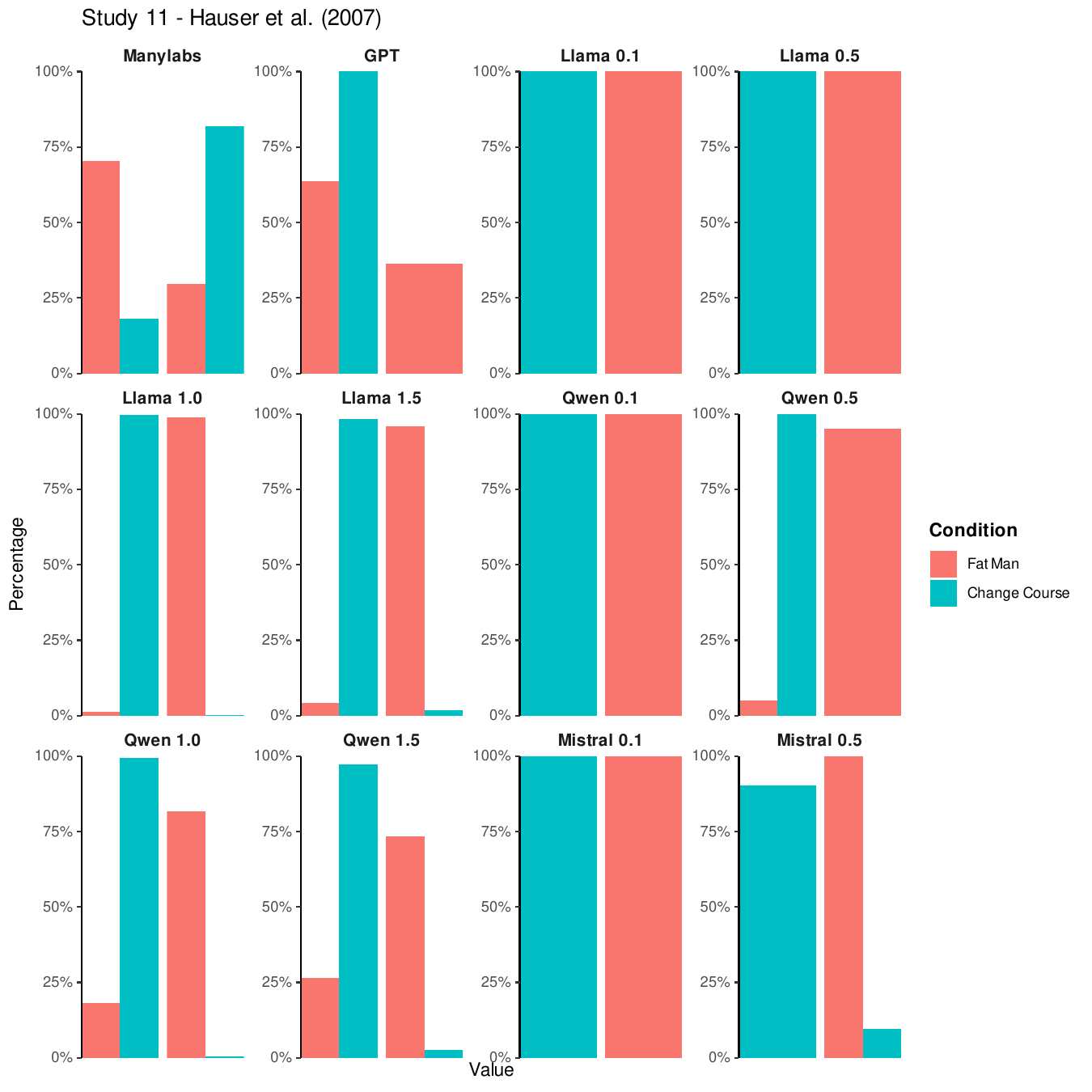}
\end{figure*}

\begin{figure*}[!tb]
\includegraphics[width=\linewidth]{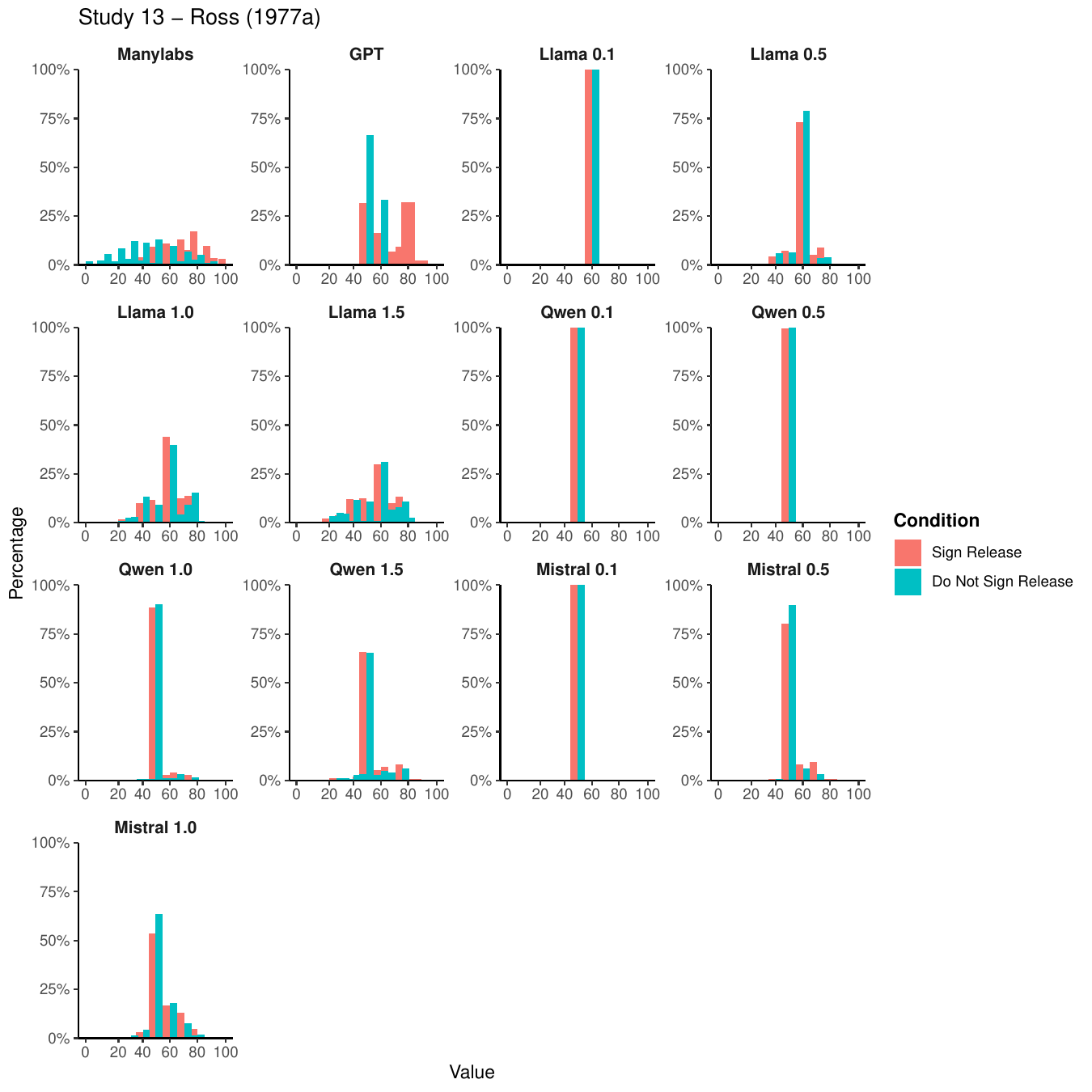}
\end{figure*}

\begin{figure*}[!tb]
\includegraphics[width=\linewidth]{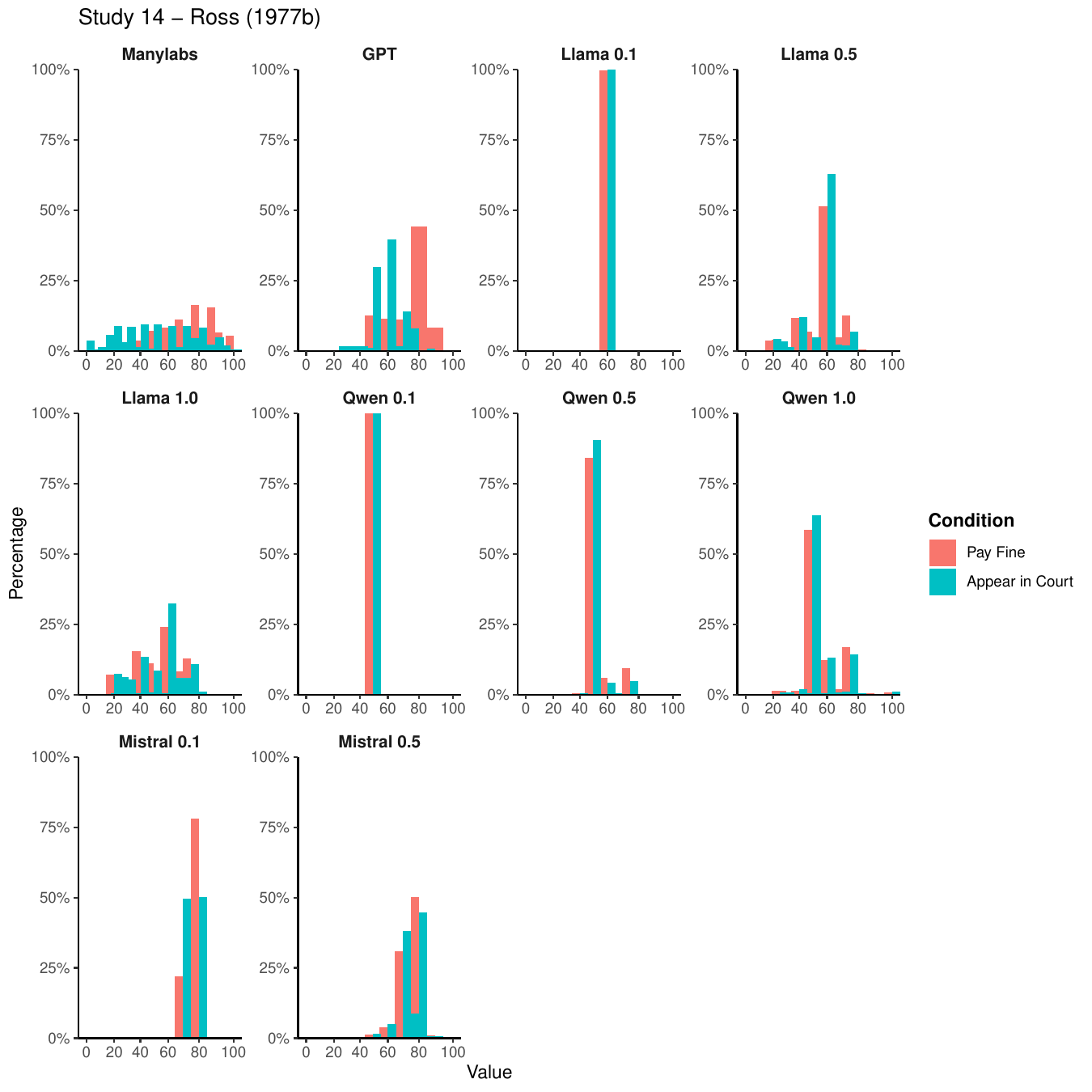}
\end{figure*}

\begin{figure*}[!tb]
\includegraphics[width=\linewidth]{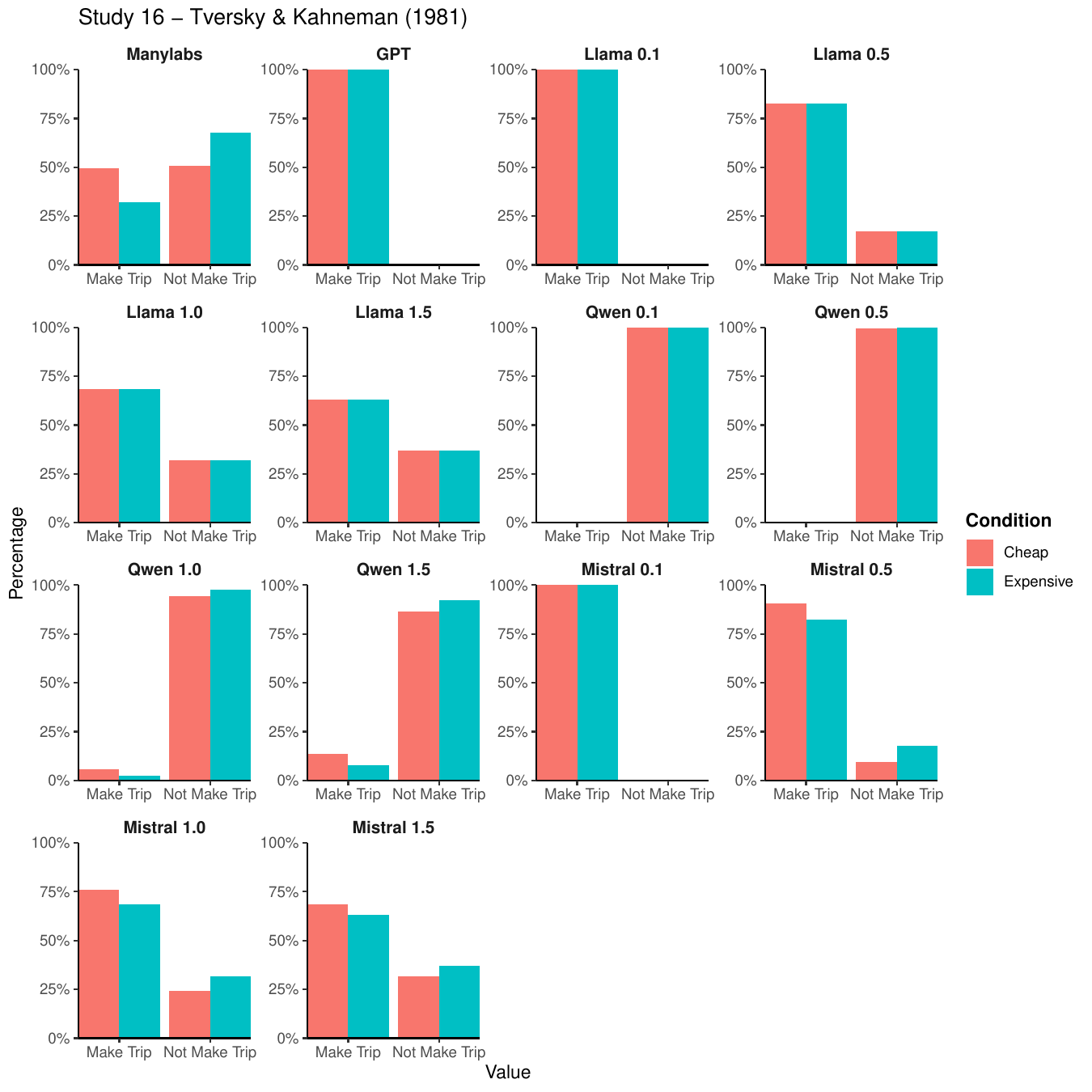}
\end{figure*}

\begin{figure*}[!tb]
\includegraphics[width=\linewidth]{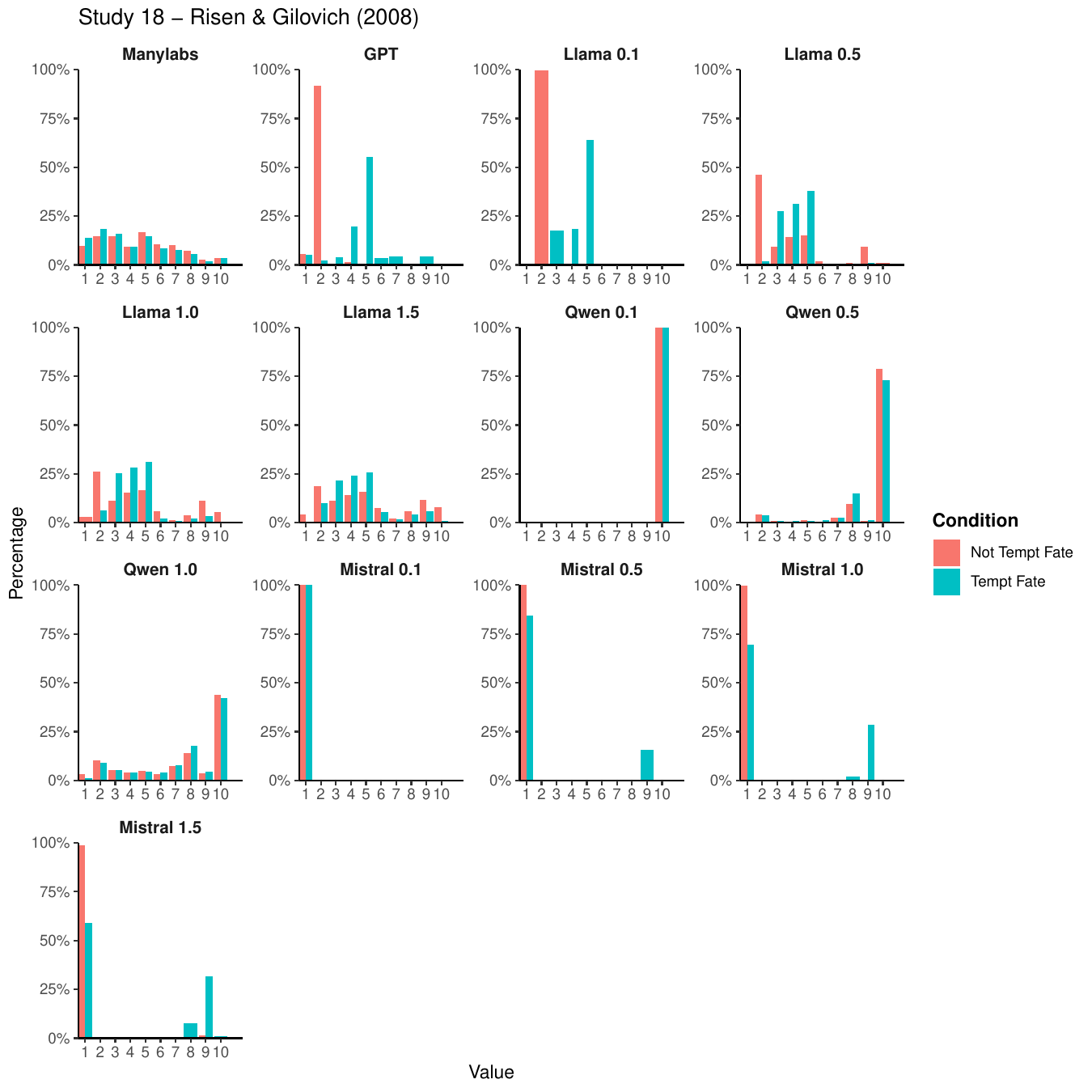}
\end{figure*}

\begin{figure*}[!tb]
\includegraphics[width=\linewidth]{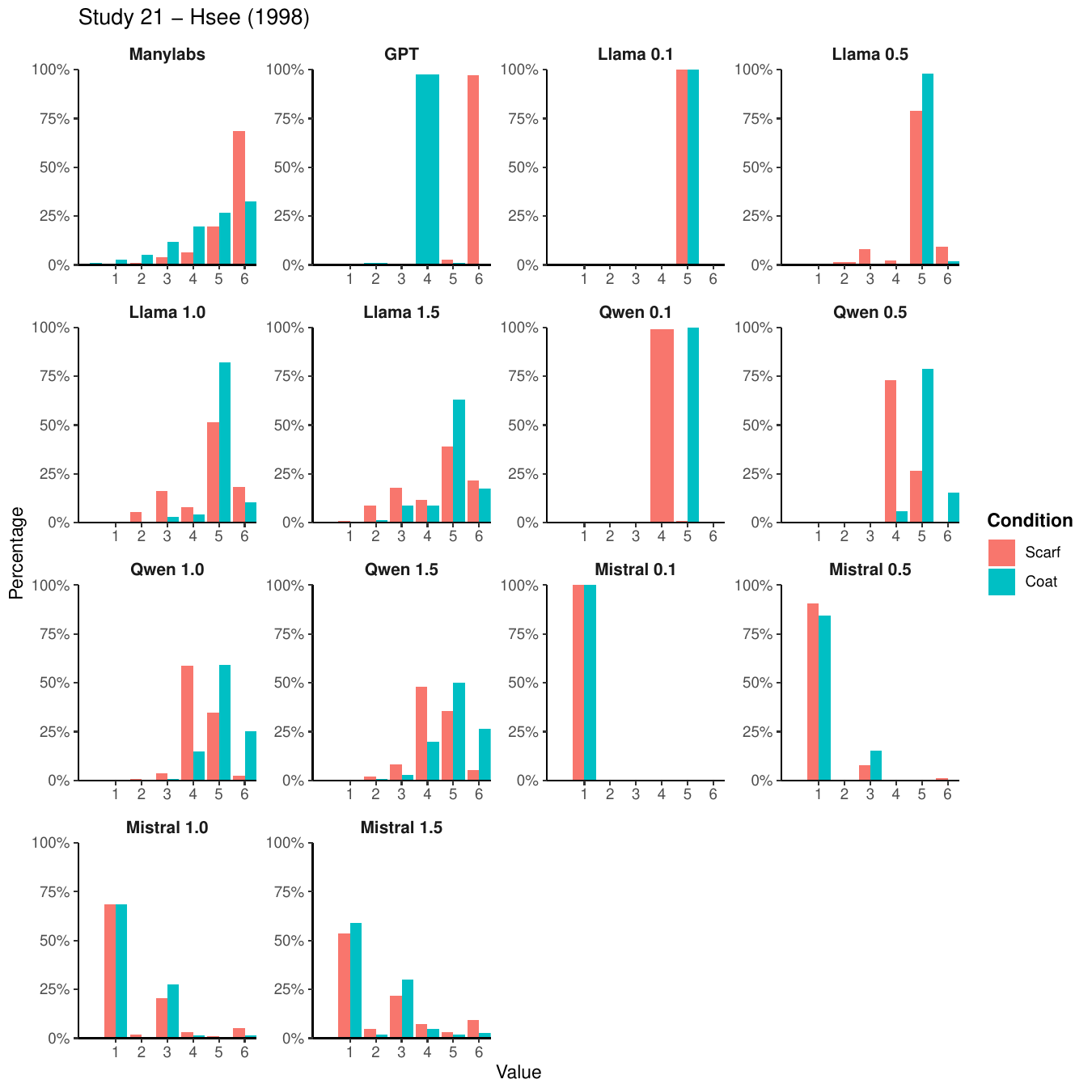}
\end{figure*}

\begin{figure*}[!tb]
\includegraphics[width=\linewidth]{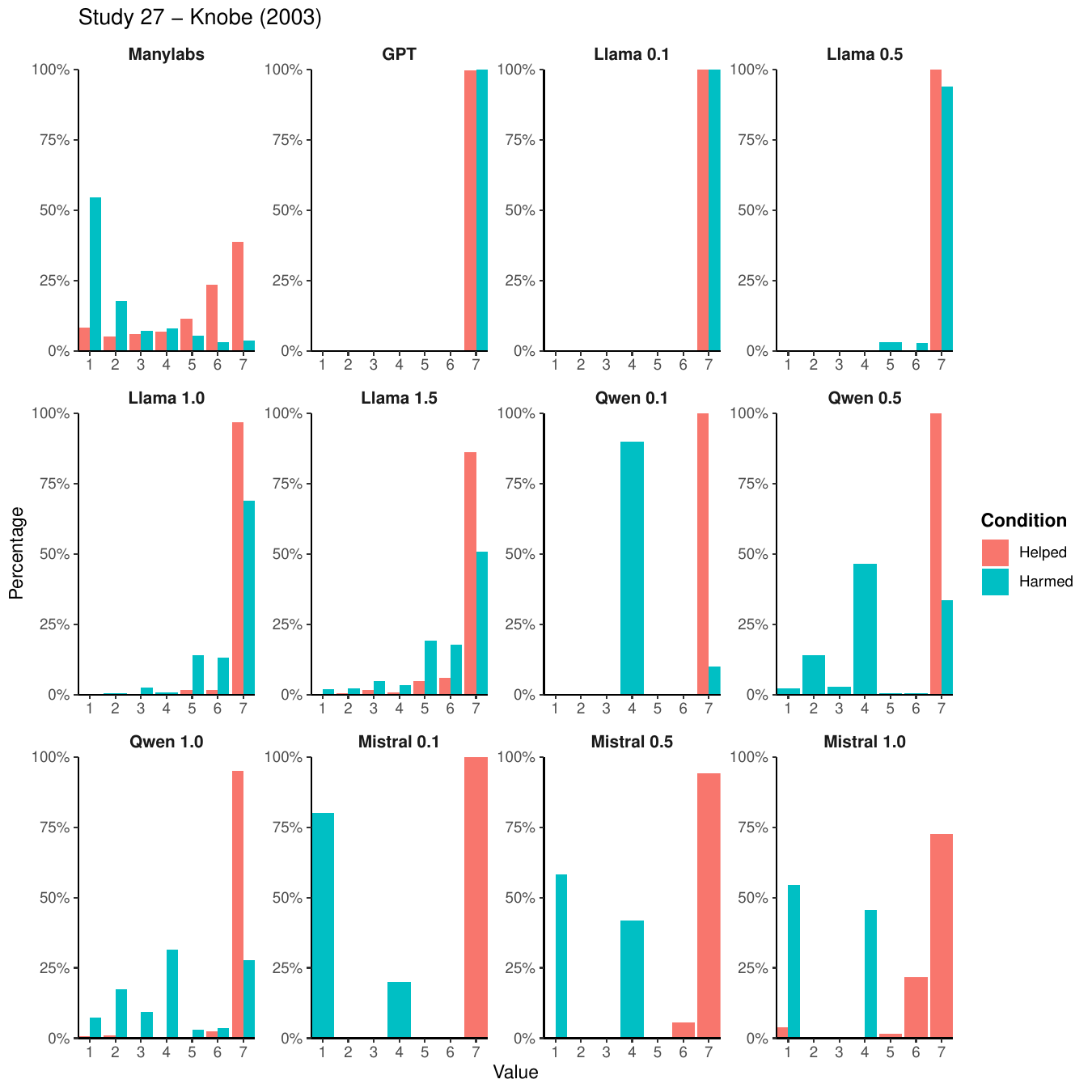}
\end{figure*}

\clearpage

\section{Effect size with temperature}
\label{app:eff_size}
\begin{figure*}[!tb]
\includegraphics[width=\linewidth]{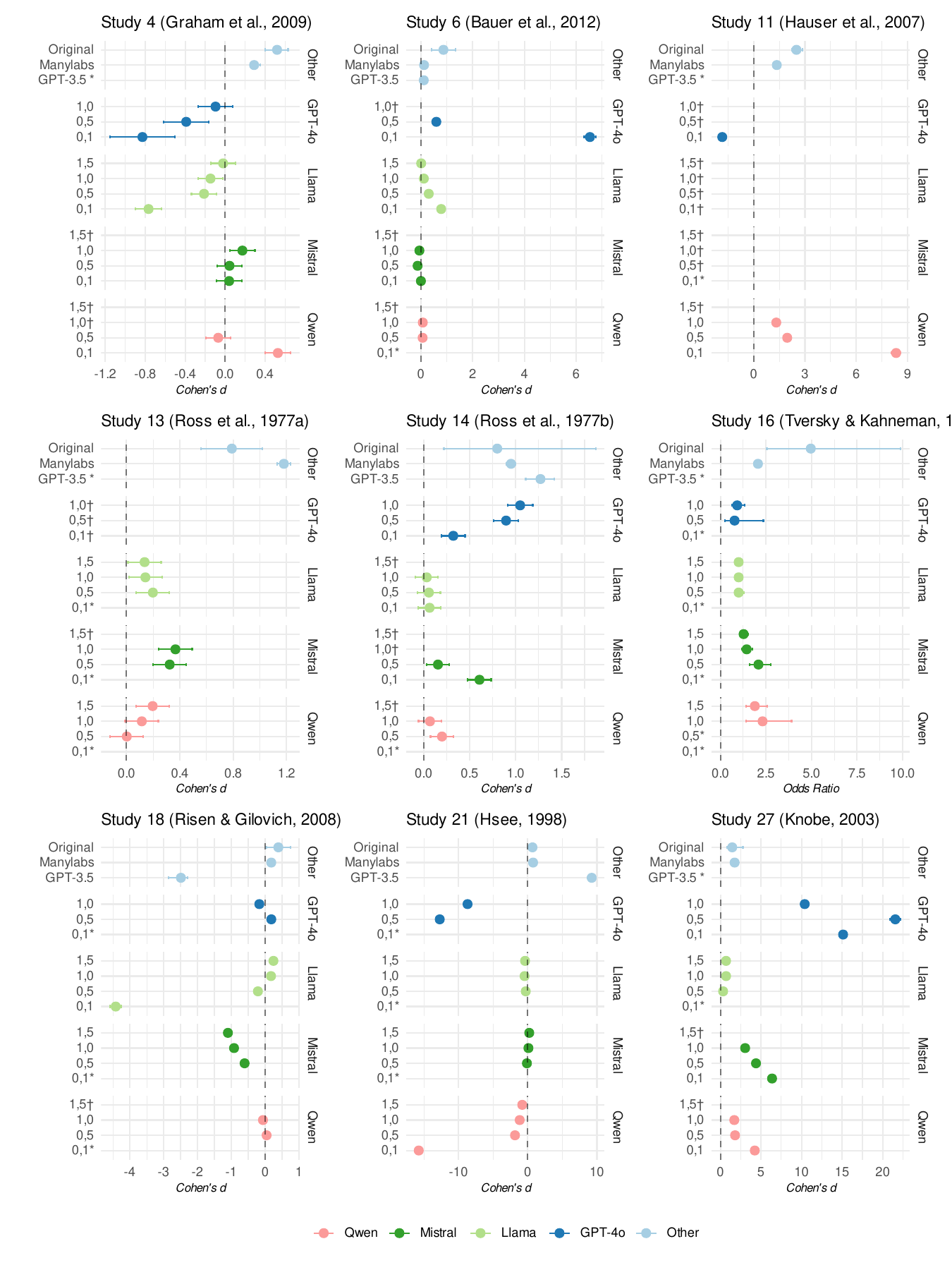}
\caption{Effect sizes for the studies successfully replicated by many labs. The left y-axis indicates the different temperatures used. A $*$ indicates that no effect could be calculated due to a correct response effect, while a $\dagger$ indicates that no effect could be calculated because the LLM did not produce useful data.}
\label{fig:cohens_d_replicable} 
\end{figure*}

\begin{figure*}[!tb]
\includegraphics[width=\linewidth]{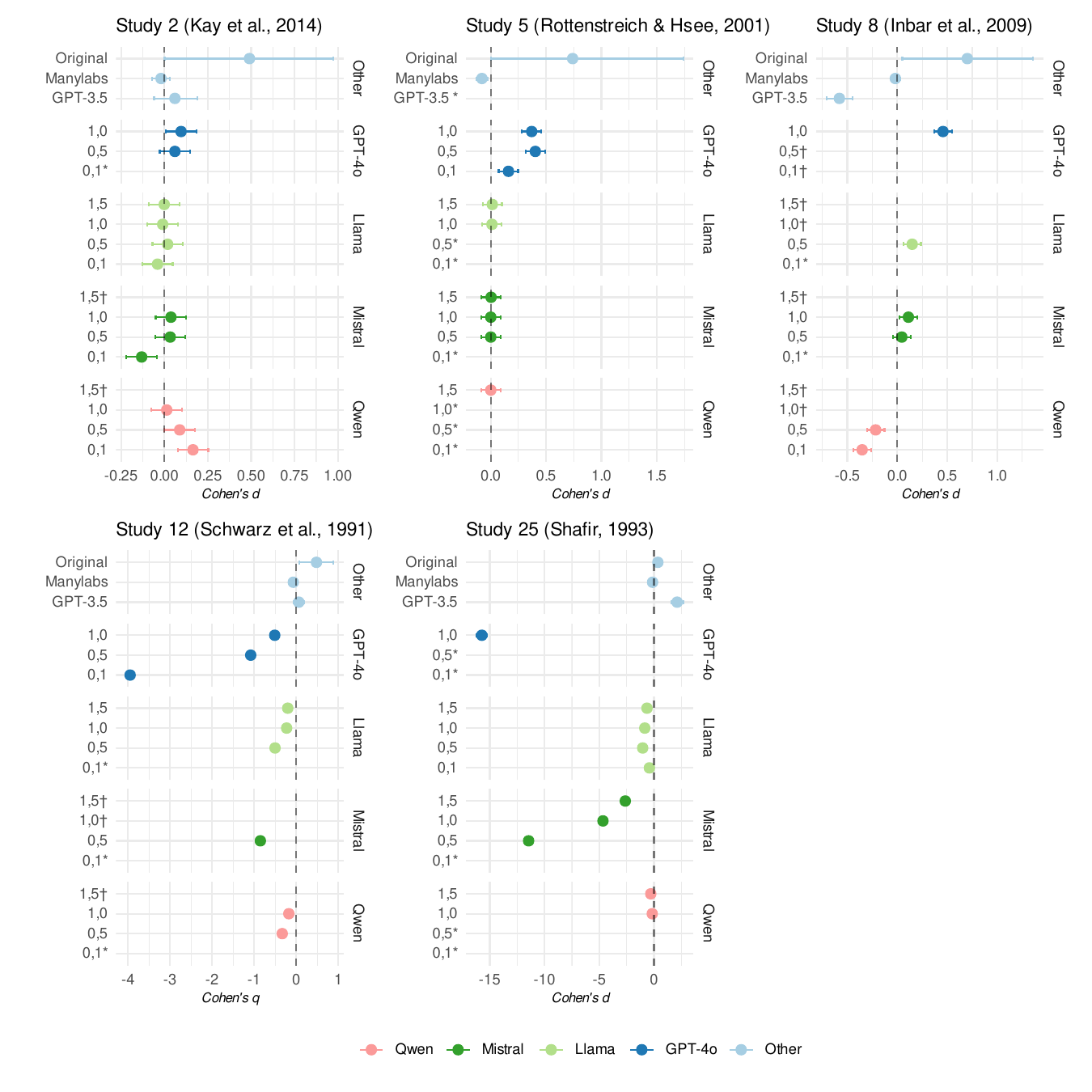}
\caption{Effect sizes for the studies unsuccessfully replicated by many labs. The left y-axis indicates the different temperatures used. A $*$ indicates that no effect could be calculated due to a correct answer effect, while a $\dagger$ indicates that no effect could be calculated because the LLM did not produce useful data.}
\label{fig:cohens_d_nonreplicable} 
\end{figure*}

Full temperature results with effect size estimations can be seen in \Cref{fig:cohens_d_replicable,fig:cohens_d_nonreplicable}. Effect sizes and p-values are reported in \Cref{tab:gpt,tab:llama,tab:qwen,tab:mist}.

\begin{table*}[!tb]
\centering
\begin{tabular}{llll}
temperature & study id & effect size &  \\
0.1 & Kay\_2013 & - & - \\
0.1 & Graham\_2009 & -0.83 & 8.50E-06 \\
0.1 & Rottenstreich\_2001 & 0.16 & 3.60E-04 \\
0.1 & Bauer\_2012 & 6.53 & 0.00E+00 \\
0.1 & Hauser\_2007 & -1.84 & 1.10E-268 \\
0.1 & Ross\_1977b & 0.32 & 5.50E-05 \\
0.1 & Tversky\_1981 & - & - \\
0.1 & Risen\_2008 & - & - \\
0.1 & Hsee\_1998 & - & - \\
0.1 & Schwarz\_1991 & -3.95 & 0.00E+00 \\
0.1 & Shafir\_1993 & - & - \\
0.1 & Knobe\_2003 & 15.12 & 0.00E+00 \\
0.5 & Kay\_2013 & 0.06 & 1.80E-01 \\
0.5 & Graham\_2009 & -0.39 & 1.30E-04 \\
0.5 & Rottenstreich\_2001 & 0.4 & 4.90E-19 \\
0.5 & Bauer\_2012 & 0.59 & 9.60E-39 \\
0.5 & Ross\_1977b & 0.9 & 2.10E-25 \\
0.5 & Tversky\_1981 & 0.78 & 8.00E-01 \\
0.5 & Risen\_2008 & 0.18 & 5.10E-05 \\
0.5 & Hsee\_1998 & -12.71 & 0.00E+00 \\
0.5 & Schwarz\_1991 & -1.08 & 0.00E+00 \\
0.5 & Shafir\_1993 & - & - \\
0.5 & Knobe\_2003 & 21.55 & 0.00E+00 \\
1 & Kay\_2013 & 0.1 & 3.30E-02 \\
1 & Graham\_2009 & -0.1 & 2.20E-01 \\
1 & Rottenstreich\_2001 & 0.37 & 2.50E-16 \\
1 & Inbar\_2009 & 0.46 & 6.40E-24 \\
1 & Ross\_1977b & 1.05 & 4.00E-32 \\
1 & Tversky\_1981 & 0.92 & 7.10E-01 \\
1 & Risen\_2008 & -0.17 & 1.20E-04 \\
1 & Hsee\_1998 & -8.69 & 0.00E+00 \\
1 & Schwarz\_1991 & -0.51 & 0.00E+00 \\
1 & Shafir\_1993 & -15.73 & 7.90E-01 \\
1 & Knobe\_2003 & 10.39 & 0.00E+00
\end{tabular}
\caption{\label{tab:gpt} Replication results for GPT-4o}
\end{table*}

\begin{table*}[!tb]
\centering
\begin{tabular}{rlrr}
\multicolumn{1}{l}{temperature} & study id & \multicolumn{1}{l}{effect size} & \multicolumn{1}{l}{p-value} \\
0.1 & Kay\_2013 & -0.04 & 3.71E-01 \\
0.1 & Graham\_2009 & -0.77 & 3.62E-32 \\
0.1 & Rottenstreich\_2001 & \multicolumn{1}{l}{-} & \multicolumn{1}{l}{-} \\
0.1 & Bauer\_2012 & 0.78 & 4.93E-64 \\
0.1 & Inbar\_2009 & \multicolumn{1}{l}{-} & \multicolumn{1}{l}{-} \\
0.1 & Ross\_1977a & \multicolumn{1}{l}{-} & \multicolumn{1}{l}{-} \\
0.1 & Ross\_1977b & 0.06 & 3.18E-01 \\
0.1 & Tversky\_1981 & \multicolumn{1}{l}{-} & \multicolumn{1}{l}{-} \\
0.1 & Risen\_2008 & -4.42 & 0.00E+00 \\
0.1 & Hsee\_1998 & \multicolumn{1}{l}{-} & \multicolumn{1}{l}{-} \\
0.1 & Schwarz\_1991 & \multicolumn{1}{l}{-} & \multicolumn{1}{l}{-} \\
0.1 & Shafir\_1993 & -0.42 & 0.00E+00 \\
0.1 & Knobe\_2003 & \multicolumn{1}{l}{-} & \multicolumn{1}{l}{-} \\
0.5 & Kay\_2013 & 0.02 & 6.86E-01 \\
0.5 & Graham\_2009 & -0.21 & 1.16E-03 \\
0.5 & Rottenstreich\_2001 & \multicolumn{1}{l}{-} & \multicolumn{1}{l}{-} \\
0.5 & Bauer\_2012 & 0.30 & 1.77E-11 \\
0.5 & Inbar\_2009 & 0.15 & 8.99E-04 \\
0.5 & Ross\_1977a & 0.20 & 1.80E-03 \\
0.5 & Ross\_1977b & 0.06 & 3.68E-01 \\
0.5 & Tversky\_1981 & 1.00 & 1.00E+00 \\
0.5 & Risen\_2008 & -0.21 & 1.77E-06 \\
0.5 & Hsee\_1998 & -0.28 & 4.05E-10 \\
0.5 & Schwarz\_1991 & -0.50 & 0.00E+00 \\
0.5 & Shafir\_1993 & -1.04 & 0.00E+00 \\
0.5 & Knobe\_2003 & 0.33 & 1.73E-13 \\
1 & Kay\_2013 & -0.01 & 8.15E-01 \\
1 & Graham\_2009 & -0.15 & 2.30E-02 \\
1 & Rottenstreich\_2001 & 0.01 & 8.18E-01 \\
1 & Bauer\_2012 & 0.11 & 1.15E-02 \\
1 & Ross\_1977a & 0.14 & 2.48E-02 \\
1 & Ross\_1977b & 0.03 & 5.93E-01 \\
1 & Tversky\_1981 & 1.00 & 1.00E+00 \\
1 & Risen\_2008 & 0.17 & 1.17E-04 \\
1 & Hsee\_1998 & -0.45 & 2.74E-23 \\
1 & Schwarz\_1991 & -0.23 & 2.97E-07 \\
1 & Shafir\_1993 & -0.84 & 0.00E+00 \\
1 & Knobe\_2003 & 0.70 & 1.40E-51 \\
1.5 & Kay\_2013 & 0.00 & 9.74E-01 \\
1.5 & Graham\_2009 & -0.02 & 7.52E-01 \\
1.5 & Rottenstreich\_2001 & 0.01 & 7.59E-01 \\
1.5 & Bauer\_2012 & 0.01 & 8.38E-01 \\
1.5 & Ross\_1977a & 0.14 & 3.17E-02 \\
1.5 & Tversky\_1981 & 1.00 & 1.00E+00 \\
1.5 & Risen\_2008 & 0.24 & 5.19E-08 \\
1.5 & Hsee\_1998 & -0.38 & 3.24E-17 \\
1.5 & Schwarz\_1991 & -0.20 & 6.23E-06 \\
1.5 & Shafir\_1993 & -0.64 & 0.00E+00 \\
1.5 & Knobe\_2003 & 0.69 & 4.67E-51
\end{tabular}
\caption{\label{tab:llama} Replication results for Llama}
\end{table*}

\begin{table*}[!tb]
\centering
\begin{tabular}{llrr}
\toprule
\multicolumn{1}{l}{Temperature} & Study ID & \multicolumn{1}{l}{Effect Size} & \multicolumn{1}{l}{p-Value} \\
\midrule
0.1 & Kay\_2013 & 0.16 & 2.47E-04 \\
0.1 & Graham\_2009 & 0.53 & 1.54E-18 \\
0.1 & Rottenstreich\_2001 & \multicolumn{1}{l}{-} & \multicolumn{1}{l}{-} \\
0.1 & Bauer\_2012 & \multicolumn{1}{l}{-} & \multicolumn{1}{l}{-} \\
0.1 & Inbar\_2009 & -0.35 & 5.56E-15 \\
0.1 & Hauser\_2007 & 8.33 & 0.00E+00 \\
0.1 & Ross\_1977a & \multicolumn{1}{l}{-} & \multicolumn{1}{l}{-} \\
0.1 & Ross\_1977b & \multicolumn{1}{l}{-} & \multicolumn{1}{l}{-} \\
0.1 & Tversky\_1981 & \multicolumn{1}{l}{-} & \multicolumn{1}{l}{-} \\
0.1 & Risen\_2008 & \multicolumn{1}{l}{-} & \multicolumn{1}{l}{-} \\
0.1 & Hsee\_1998 & -15.74 & 0.00E+00 \\
0.1 & Schwarz\_1991 & \multicolumn{1}{l}{-} & \multicolumn{1}{l}{-} \\
0.5 & Shafir\_1993 & \multicolumn{1}{l}{-} & 0.00E+00 \\
0.5 & Knobe\_2003 & 4.23 & 0.00E+00 \\
0.5 & Kay\_2013 & 0.09 & 5.03E-02 \\
0.5 & Graham\_2009 & -0.07 & 2.69E-01 \\
0.5 & Rottenstreich\_2001 & \multicolumn{1}{l}{-} & \multicolumn{1}{l}{-} \\
0.5 & Bauer\_2012 & 0.06 & 1.48E-01 \\
0.5 & Inbar\_2009 & -0.22 & 1.30E-06 \\
0.5 & Hauser\_2007 & 1.96 & 2.43E-294 \\
0.5 & Ross\_1977a & 0.00 & 9.71E-01 \\
0.5 & Ross\_1977b & 0.20 & 1.77E-03 \\
0.5 & Tversky\_1981 & \multicolumn{1}{l}{-} & \multicolumn{1}{l}{-} \\
0.5 & Risen\_2008 & 0.04 & 3.42E-01 \\
0.5 & Hsee\_1998 & -1.84 & 1.40E-267 \\
0.5 & Schwarz\_1991 & -0.33 & 9.24E-14 \\
0.5 & Shafir\_1993 & \multicolumn{1}{l}{-} & \multicolumn{1}{l}{-} \\
0.5 & Knobe\_2003 & 1.80 & 1.32E-260 \\
1 & Kay\_2013 & 0.01 & 7.73E-01 \\
1 & Rottenstreich\_2001 & \multicolumn{1}{l}{-} & \multicolumn{1}{l}{-} \\
1 & Bauer\_2012 & 0.07 & 1.05E-01 \\
1 & Hauser\_2007 & 1.32 & 4.78E-159 \\
1 & Ross\_1977a & 0.11 & 7.10E-02 \\
1 & Ross\_1977b & 0.07 & 2.77E-01 \\
1 & Tversky\_1981 & 2.31 & 6.67E-04 \\
1 & Risen\_2008 & -0.06 & 1.49E-01 \\
1 & Hsee\_1998 & -1.16 & 1.99E-127 \\
1 & Schwarz\_1991 & -0.18 & 7.89E-05 \\
1 & Shafir\_1993 & -0.16 & 0.00E+00 \\
1 & Knobe\_2003 & 1.70 & 3.87E-239 \\
1.5 & Rottenstreich\_2001 & 0.00 & 1.00E+00 \\
1.5 & Ross\_1977a & 0.20 & 1.96E-03 \\
1.5 & Tversky\_1981 & 1.89 & 2.62E-05 \\
1.5 & Hsee\_1998 & -0.79 & 1.36E-64 \\
1.5 & Shafir\_1993 & -0.30 & 0.00E+00\\
\bottomrule
\end{tabular}
\caption{\label{tab:qwen} Replication results for Qwen}
\end{table*}

\begin{table*}[!tb]
\centering
\begin{tabular}{llrr}
\toprule
Temperature &Study ID & Effect Size & p-Value \\
\midrule
0.1 & Kay\_2013 & -0.13 & 3.26E-03 \\
0.1 & Graham\_2009 & 0.04 & 4.81E-01 \\
0.1 & Rottenstreich\_2001 & - & - \\
0.1 & Bauer\_2012 & 0 & 1.00E+00 \\
0.1 & Inbar\_2009 & - & - \\
0.1 & Hauser\_2007 & - & - \\
0.1 & Ross\_1977a & - & - \\
0.1 & Ross\_1977b & 0.61 & 4.64E-21 \\
0.1 & Tversky\_1981 & - & - \\
0.1 & Risen\_2008 & - & - \\
0.1 & Hsee\_1998 & - & - \\
0.1 & Schwarz\_1991 & - & - \\
0.1 & Shafir\_1993 & - & - \\
0.1 & Knobe\_2003 & 6.36 & 0.00E+00 \\
0.5 & Kay\_2013 & 0.03 & 4.60E-01 \\
0.5 & Graham\_2009 & 0.04 & 4.62E-01 \\
0.5 & Rottenstreich\_2001 & 0 & 1.00E+00 \\
0.5 & Bauer\_2012 & -0.13 & 4.12E-03 \\
0.5 & Inbar\_2009 & 0.04 & 3.17E-01 \\
0.5 & Ross\_1977a & 0.32 & 5.08E-07 \\
0.5 & Ross\_1977b & 0.16 & 1.40E-02 \\
0.5 & Tversky\_1981 & 2.09 & 6.16E-08 \\
0.5 & Risen\_2008 & -0.61 & 2.75E-40 \\
0.5 & Hsee\_1998 & -0.11 & 1.16E-02 \\
0.5 & Schwarz\_1991 & -0.85 & 0.00E+00 \\
0.5 & Shafir\_1993 & -11.45 & 5.02E-01 \\
0.5 & Knobe\_2003 & 4.38 & 0.00E+00 \\
1 & Kay\_2013 & 0.04 & 4.06E-01 \\
1 & Graham\_2009 & 0.17 & 5.46E-03 \\
1 & Rottenstreich\_2001 & 0 & 1.00E+00 \\
1 & Bauer\_2012 & -0.06 & 1.88E-01 \\
1 & Inbar\_2009 & 0.11 & 1.34E-02 \\
1 & Ross\_1977a & 0.37 & 7.38E-09 \\
1 & Tversky\_1981 & 1.44 & 3.29E-04 \\
1 & Risen\_2008 & -0.92 & 2.09E-85 \\
1 & Hsee\_1998 & 0.1 & 1.90E-02 \\
1 & Shafir\_1993 & -4.66 & 2.73E-03 \\
1 & Knobe\_2003 & 3.04 & 0.00E+00 \\
1.5 & Rottenstreich\_2001 & 0 & 9.53E-01 \\
1.5 & Tversky\_1981 & 1.27 & 1.25E-02 \\
1.5 & Risen\_2008 & -1.1 & 4.42E-117 \\
1.5 & Hsee\_1998 & 0.22 & 5.40E-07 \\
1.5 & Shafir\_1993 & -2.62 & 1.71E-06\\
\bottomrule
\end{tabular}
\caption{\label{tab:mist} Replication results for Mistral}
\end{table*}

\end{document}